\documentclass{article}

\usepackage{PRIMEarxiv}

\usepackage[utf8]{inputenc} % allow utf-8 input
\usepackage[T1]{fontenc}    % use 8-bit T1 fonts
\usepackage{multirow}%
\usepackage{amsmath,amssymb,amsfonts}%
\usepackage{amsthm}%
\usepackage{mathrsfs}%
\usepackage[title]{appendix}%
\usepackage{xcolor}%
\usepackage{textcomp}%
\usepackage{manyfoot}%
\usepackage{booktabs}%
\usepackage{algorithm}%
\usepackage{algorithmicx}%
\usepackage{algpseudocode}%
\usepackage{listings}%

\usepackage{hyperref}
\usepackage{tabularx}
\usepackage{longtable}
\usepackage{enumitem}
\usepackage{colortbl}
\usepackage{color,soul}
\usepackage{lmodern,textcomp}
\definecolor{green}{HTML}{47a63a}

\usepackage{mathtools}
\usepackage{tikz}
\usepackage{etoolbox}
\AtBeginEnvironment{tcolorbox}{\small}
\usepackage[most]{tcolorbox}
\usepackage{tikz}
\usepackage{float}

\usepackage{varwidth}
\usepackage{capt-of}
\usepackage[export]{adjustbox}
\usepackage[frozencache,cachedir=.]{minted}
\usepackage{subcaption}

\usepackage{graphicx}%
\usepackage{verbatim}
\usepackage{outlines}
\usepackage{cleveref}
\usepackage{epigraph}
\usepackage{inconsolata}

\definecolor{lightgreen}{rgb}{0.56, 0.93, 0.56}
\definecolor{lightgray}{RGB}{233, 234, 236}
\definecolor{moonstoneblue}{rgb}{0.5, 0.66, 0.76}
\definecolor{white}{RGB}{255,255,255}
\usepackage[numbers]{natbib} % Si usas números en tus citas
\usetikzlibrary{shapes.geometric}

\usepackage[group-separator={,}]{siunitx}
\usepackage{fontawesome}
\makeatletter
\newcommand\mathcircled[1]{%
  \mathpalette\@mathcircled{#1}%
}
\newcommand\@mathcircled[2]{%
  \tikz[baseline=(math.base)] \node[draw,circle,inner sep=3pt] (math) {$\m@th#1#2$};%
}
\makeatother
\setul{0.5ex}{0.3ex}

\tikzset{
    buffer/.style={
        draw,
        shape border rotate=-30,
        isosceles triangle,
        isosceles triangle apex angle=85,
        fill=black,
        node distance=0.1em,
        scale=0.1
    }
}

\newtcbox{\badge}[1][red]{
  on line, 
  arc=4pt,
  colback=#1!10!green,
  colframe=#1!10!green,
  fontupper=\color{white},
  boxrule=1pt, 
  boxsep=0pt,
  left=3pt,
  right=3pt,
  top=1pt,
  bottom=1pt
}

\newcommand{\visgraphvar}{\texttt{VisGraphVar}}

\title{\visgraphvar: A Benchmark Generator for Assessing Variability in Graph Analysis Using Large Vision-Language Models
}
\author{
  Camilo Chacón Sartori, Christian Blum, Filippo Bistaffa \\
  Artificial Intelligence Research Institute (IIIA-CSIC) \\
  Bellaterra, Spain \\
  \texttt{\{cchacon, christian.blum, filippo.bistaffa\}@iiia.csic.es}
}

\begin{document}
\maketitle

\begin{abstract}
The fast advancement of Large Vision-Language Models (LVLMs) has shown immense potential. These models are increasingly capable of tackling abstract visual tasks. Geometric structures, particularly graphs with their inherent flexibility and complexity, serve as an excellent benchmark for evaluating these models' predictive capabilities. While human observers can readily identify subtle visual details and perform accurate analyses, our investigation reveals that state-of-the-art LVLMs exhibit consistent limitations in specific visual graph scenarios, especially when confronted with stylistic variations. In response to these challenges, we introduce \visgraphvar\ (\textbf{Vis}ual \textbf{Graph} \textbf{Var}iability), a customizable benchmark generator able to produce graph images for seven distinct task categories (detection, classification, segmentation, pattern recognition, link prediction, reasoning, matching), designed to systematically evaluate the strengths and limitations of individual LVLMs. We use \visgraphvar\ to produce 990 graph images and evaluate six LVLMs, employing two distinct prompting strategies, namely zero-shot and chain-of-thought. The findings demonstrate that variations in visual attributes of images (e.g., node labeling and layout) and the deliberate inclusion of visual imperfections, such as overlapping nodes, significantly affect model performance. This research emphasizes the importance of a comprehensive evaluation across graph-related tasks, extending beyond reasoning alone. \visgraphvar\ offers valuable insights to guide the development of more reliable and robust systems capable of performing advanced visual graph analysis.

\includegraphics[width=0.03\linewidth, height=0.017\textheight]{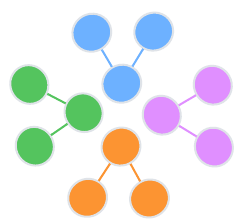} Home page: \url{https://camilochs.github.io/visgraphvar-website}

\end{abstract}

% keywords can be removed

\section{Introduction}

\epigraph{Graphs stand or fall by their choice of nodes and edges.}{---\textup{ Watts and Strogatz}
}

To prove their relevance, new technologies must face challenging scenarios. Such trials reveal their true potential and distinguish meaningful innovations from superficial hype, often lacking technical depth. The rise of foundation models presents an unprecedented opportunity to test this principle. Large Vision-Language Models (LVLMs) significantly extend the capabilities of foundation models~\cite{zhang2024visionlanguagemodelsvisiontasks}, enabling more advanced multimodal interactions and applications~\cite{yin2024surveymultimodallargelanguage}. Building on the progress of Large Language Models (LLMs)---including Transformer-based architectures like GPT~\cite{openai2024gpt4technicalreport} and BERT~\cite{devlin2019bertpretrainingdeepbidirectional}, as well as hybrid systems like BLIP~\cite{li2022blipbootstrappinglanguageimagepretraining} and Florence~\cite{yuan2021florencenewfoundationmodel}---LVLMs seamlessly integrate visual analysis into problem-solving, addressing a wide range of image-based tasks with notable effectiveness. \\

Among the tasks where these models have demonstrated to be effective are open-vocabulary object detection~\cite{jin2024llmsmeetvlmsboost},  segmentation~\cite{wang2024llmsegbridgingimagesegmentation}, classification~\cite{zhang2024visuallygroundedlanguagemodelsbad}, image captioning~\cite{dong2024benchmarkingimprovingimagecaption}, and visual question answering~\cite{wang2024surgicallvlmlearningadaptlarge}. Until now, the field responsible for investigating these tasks has been computer vision, where researchers focus on interpreting and understanding images through computational means. LVLMs come to complement this. They are capable of understanding not only images due to the massive scale of information in their training set but also text. An LVLM can receive one or more images along with a prompt, posing a task to the model and guiding its reasoning. \\

Among the leading LVLMs currently available are GPT-4o~\cite{wu2024gpt4ovisualperceptionperformance}, Claude-3.5-Sonnet~\cite{anthropicIntroducingClaude}, Gemini-Pro-1.5~\cite{geminiteam2024gemini15unlockingmultimodal}, Llama3.2~\cite{dubey2024llama3herdmodels}, Qwen-2-VL~\cite{yang2024qwen2technicalreport}, and Pixtral~\cite{agrawal2024pixtral12b}. These models can perform diverse tasks such as generating detailed descriptions of family photographs, analyzing landscapes, identifying books from cover images, and even tackling mathematical problems presented in a visual format. However, different LVLMs demonstrate varying levels of performance across different visual tasks, and as demonstrated in the works of ~\citet{li2023evaluatingobjecthallucinationlarge} and ~\citet{liu2024surveyhallucinationlargevisionlanguage}, they remain susceptible to hallucination. \\

One particularly challenging domain for LVLMs involves the analysis of geometric structures, especially graphs. The inherent complexity of graphs stems from their flexible nature, where nodes and edges can be arranged in countless configurations, creating potentially intricate patterns. The ability to effectively process graph images would be transformative, enabling LVLMs to address complex problems in graph theory across numerous applications. These include tracking information propagation in social networks, analyzing communication flows, studying biological metabolic networks, optimizing route planning for robotics and autonomous vehicles, facilitating circuit design, and modeling dynamic simulations. This widespread applicability demonstrates how graphs serve as fundamental tools across diverse practical domains~\cite{hegeman2018surveygraphanalysisapplications}. \\

\noindent Our research seeks to address the following two fundamental questions: 
\begin{enumerate}[topsep=5pt]
    \item Primary question: How robust are LVLMs in analyzing visual graphs?
    \item Secondary question: To what extent does the choice of visual style---such as node color, node labels, or layout---impact LVLM performance?
\end{enumerate} 
To investigate these, we developed \visgraphvar\ (\textbf{Vis}ual \textbf{Graph} \textbf{Var}iability), a visual graph benchmark generator capable of producing graph images with varying styles and structures. We developed \visgraphvar\ under the assumption that a benchmark is only good if, for its generation, lessons learned from previous benchmarks are taken into account. While existing benchmarks for graph image analysis often focus on single tasks like reasoning~\cite{wei2024gitagraphvisualtextual} and overlook style variations~\cite{Williams2024-rp,li2024visiongraphleveraginglargemultimodal}, \visgraphvar\ operates across multiple dimensions. Inspired by the work of~\cite{Chartrand2012-iz}, we divide our analysis into seven dimensions, each examining a specific task related to visual graph interpretation:

\begin{enumerate}[topsep=5pt]
    \item \textbf{Node and Edge Detection.} Question: How many nodes and edges does the graph contain?
    \item \textbf{Graph Type Classification.} Question: What type of graph is shown in the image?
    \item \textbf{Segmentation.} Question: Which edges of the graph are cut-edges?
    \item \textbf{Pattern Recognition.} Question: Which patterns can be identified in the graph, and how many instances of each?
    \item \textbf{Link Prediction.} Question: Based on the graph's structure, which nodes are likely to form new connections?
    \item \textbf{Reasoning.} Question: Which one is the shortest path between two specified nodes?
    \item \textbf{Matching.} Question: Are the two displayed graphs equal to each other (apart from colors and styles)?
\end{enumerate}

Through this systematic approach, \visgraphvar\ provides a comprehensive evaluation methodology capable of better capturing the strengths and weaknesses of current LVLMs in visual graph analysis tasks. This multi-dimensional strategy offers a more precise insight on how LVLMs perform across different aspects of graph interpretation and understanding. 

\subsection{Contribution and Paper Outline}

Our contribution---and what distinguishes us from other benchmark generators and benchmarks---can be summarised in the following points:
\begin{itemize}[topsep=5pt]
\item \visgraphvar\ is customizable and user-friendly, enabling the creation of datasets that effectively help to identify LVLMs' weaknesses in graph vision tasks.
\item \visgraphvar\ covers a diverse set of tasks. Rather than creating a dataset focused on a single analytical task, we encompass seven distinct tasks, allowing for more comprehensive conclusions about which LVLM performs best in specific graph-related scenarios.
\item \visgraphvar\ aims for a more realistic approach by intentionally incorporating visual imperfections (such as overlapping nodes) rather than striving for idealized graph representations. This approach enables a more authentic evaluation of LVLM capabilities, particularly given that human observers can readily identify and interpret such visual complexities.
\end{itemize}

In other words, \visgraphvar\ expands the focus on spatial layout properties---including node distribution and edge placement---and visual attributes such as node coloring and labeling. We demonstrate that stylistic elements significantly influence LVLMs' inference performance through the seven distinct tasks mentioned above. This paper unfolds as follows. Section~\ref{sec:related-work} situates our work within the existing literature. We then present a comprehensive examination of \visgraphvar's tasks and the benchmark generated for this paper in Section~\ref{sec:VisGraphVar}. Section~\ref{experiments} offers a thorough evaluation of six LVLMs through both quantitative and qualitative analyses. The broader implications and emerging research questions are discussed in Section~\ref{sec:discussion}. We conclude in Section~\ref{sec:conclu} with a summary of our findings and potential avenues for future research. The results from Section~\ref{experiments} show that both stylistic variations and the different tasks introduced by \visgraphvar\ impact model performance. This indicates that the benchmark presents a significant challenge, with no model able to excel across all tasks consistently. 

\section{Related work}\label{sec:related-work}

Automating the interpretation of images without dependence on labeled training data has long been a central challenge in computer vision. Traditional vision-language models like VisualBERT~\cite{li2019visualbertsimpleperformantbaseline}, VilBERT~\cite{lu2019vilbertpretrainingtaskagnosticvisiolinguistic}, and LXMERT~\cite{tan2019lxmertlearningcrossmodalityencoder} (2019-2020) required task-specific labeled datasets for operations such as Visual Question Answering and Image Captioning. CLIP (Contrastive Language Image Pretraining) changed this paradigm as the first model to enable large-scale image interpretation without task-specific labels~\cite{radford2021learningtransferablevisualmodels}. Through its dual transformer-based architecture~\cite{dosovitskiy2021imageworth16x16words, vaswani2023attentionneed}, combining image and text encoders, CLIP can interpret image-text relationships in a zero-shot manner, forming novel connections without needing prior examples. However, not all images pose the same difficulty for a model to interpret them. \\

The emergence of advanced multimodal LVLMs like Gemini Pro-1.5~\cite{geminiteam2024gemini15unlockingmultimodal}, GPT-4o~\cite{wu2024gpt4ovisualperceptionperformance}, Llama-3.2~\cite{dubey2024llama3herdmodels}, and Claude-3.5-Sonnet~\cite{anthropicIntroducingClaude} has sparked interest in testing their capabilities with complex geometric structures like graphs. Recent benchmarks evaluate these models' graph understanding abilities: VisionGraph~\cite{li2024visiongraphleveraginglargemultimodal} focuses on algorithmic tasks like cycle detection and shortest path finding using few-shot and chain-of-thought prompting. At the same time, GITA~\cite{wei2024gitagraphvisualtextual} offers an end-to-end framework that leverages visual graph representations to enable general graph reasoning. However, it is crucial to recognize that visual graph analysis encompasses a much broader scope of challenges beyond the reasoning tasks that GITA addresses. Williams et al.~\cite{Williams2024-rp} have also proposed a benchmark, though with a more limited task scope and LVLM coverage. Although DynaMath~\cite{Zou2024-tt} is not exclusively dedicated to graph evaluation, it incorporates graph analysis as one component of its assessment methodology. Nevertheless, analogous to GITA, its primary focus remains on visual reasoning tasks. All these benchmarks, though significant, examine only a narrow range of graph visualization formats, ignoring key visual aspects of 2D graph representations.

\section{\visgraphvar: A benchmark generator}\label{sec:VisGraphVar}

We present \visgraphvar, a customizable benchmark generator that produces evaluation datasets designed to assess the robustness of LVLMs in multimodal graph analysis tasks. This benchmark generator offers a novel approach to evaluating a critical aspect of graph images: \textit{the ability to interpret representational variability}. This variability is measured through seven independent graph-analysis tasks designed to test model performance. This allows us to pinpoint each model’s strengths and weaknesses across a range of different tasks, providing a comprehensive, independent assessment of LVLM performance.

\subsection{A Custom Synthetic Dataset}

\visgraphvar\ was implemented in Python 3.11 using the NetworkX library for graph image generation. The capacity for customization and extensibility constitutes a fundamental requirement in benchmark design~\cite{10.1145/2668930.2688819}. Our generator fully implements these capabilities through its modular architecture. The graph visualization pipeline supports multiple parameterization options:

\begin{enumerate}[topsep=5pt, itemsep=5pt]
    \item \textbf{Layout Selection and Its Impact.} Previous graph image benchmarks have largely overlooked the importance of graph layout, typically resorting to aesthetically pleasing arrangements or using standard library defaults (e.g., NetworkX in Python). However, since their training data influences LLVMs' inferences, the spatial arrangement of nodes and edges can significantly impact their ability to detect evident visual patterns. In general, layout selection is crucial in many graph applications because the inherent flexibility of graph structures allows us to emphasize specific characteristics over others. For example:
    \begin{itemize}[topsep=5pt]
        \item The spring layout effectively highlights nodes with high connectivity centrality
        \item A circular layout excels at visualizing disconnected components by maintaining their visibility while keeping them organized separately from the main graph
        \item A spectral layout effectively reveals community structures and node clusters by positioning nodes based on global connectivity patterns
    \end{itemize}
    The choice of layout, therefore, depends on which graph properties need emphasis~\cite{10.5555/2381255.2381265}. Consequently, if an LLVM fails to perform consistently across different layouts, its practical applications could be significantly limited. This layout sensitivity becomes particularly critical when the same graph structure, presented in different spatial arrangements, yields varying inference results.
    \item \textbf{Configurable Graph Element Parameters.} The ability to adjust node quantities and edge connection probabilities enables evaluation of how visual complexity affects LLVM performance. This configurability allows us to assess, for example, whether increasing the number of elements in the image degrades LLVM response quality.
   \item \textbf{Color Schemes and Text Labels.} Existing benchmarks often overlook the significance of color variations and label presence in graph visualizations. This oversight is particularly concerning since real-world graph applications heavily emphasize stylistic elements. The ability of LLVMs to capture these subtle visual attributes so far remains uncertain.
\end{enumerate}

Although GITA~\cite{wei2024gitagraphvisualtextual} makes use of a dataset with visual graph attributes such as layout selection, node shape, outline style, and edge thickness, the presented evaluation remains limited to reasoning tasks. \visgraphvar\ enables a more comprehensive approach by broadening the range of visual variations and structuring the analysis with multiple tasks, creating a robust evaluation framework capable of assessing both current and future LVLMs.

\subsection{Tasks}

\begin{figure}[p]
    \centering
    \includegraphics[width=0.82\linewidth]{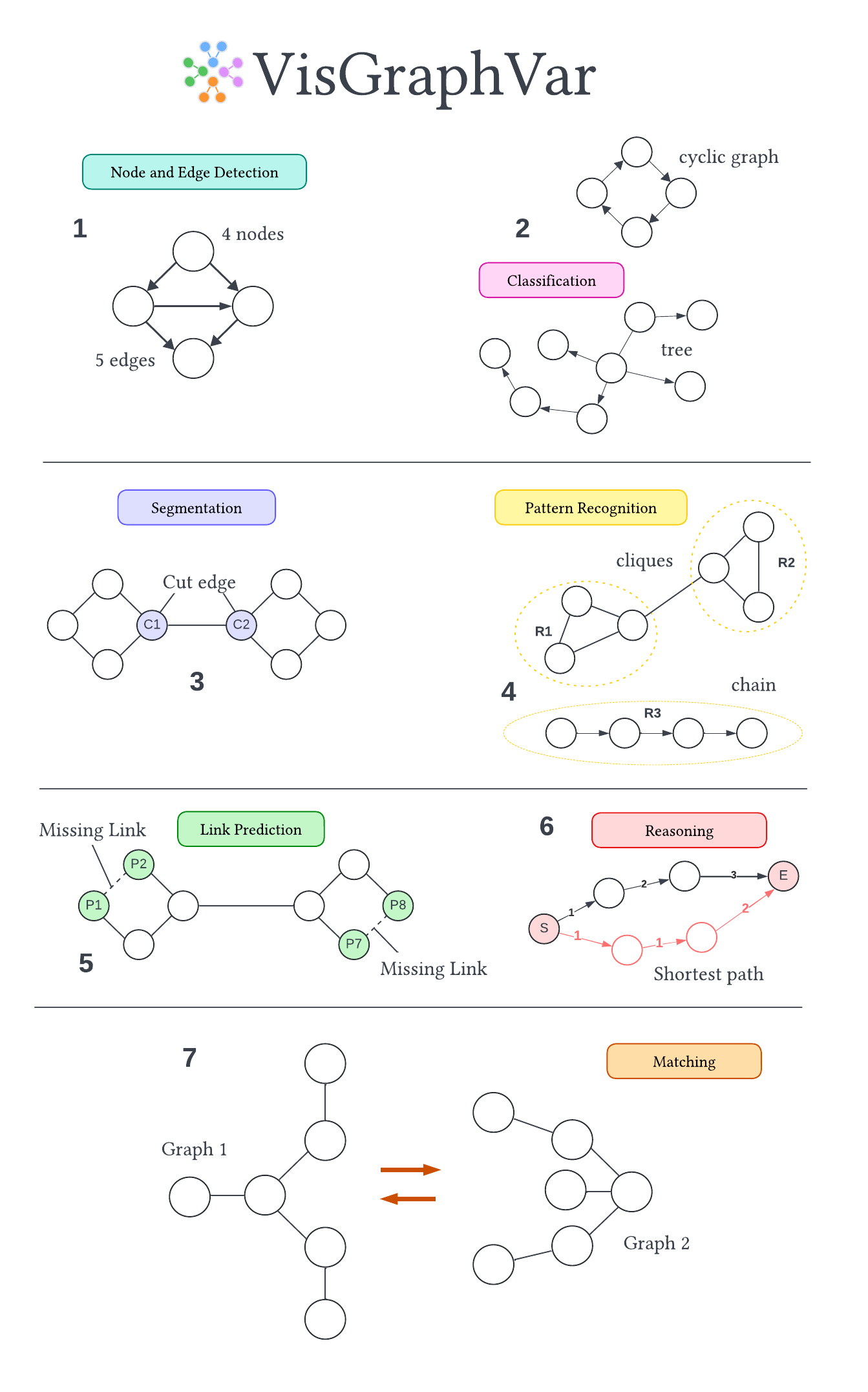}
        \caption{A general overview of the seven tasks covered by \visgraphvar\ (1-7), each representing a different challenge for LVLMs, enabling us to conduct a more detailed performance comparison and evaluation.}
    \label{fig:tasks}
\end{figure}

A benchmark that only covers limited aspects of a technology may overlook potential shortcomings. In response to this challenge, \visgraphvar\ implements seven comprehensive evaluation tasks (see Figure~\ref{fig:tasks}), each designed to assess distinct aspects of LVLM graph comprehension capabilities while maintaining compatibility with all aforementioned customization options. Below is an explanation of each task and the reasoning behind the choices made.

\subsubsection{Task 1: Node and Edge Detection}

\begin{figure}[!t]
    \centering
    \includegraphics[width=0.8\linewidth]{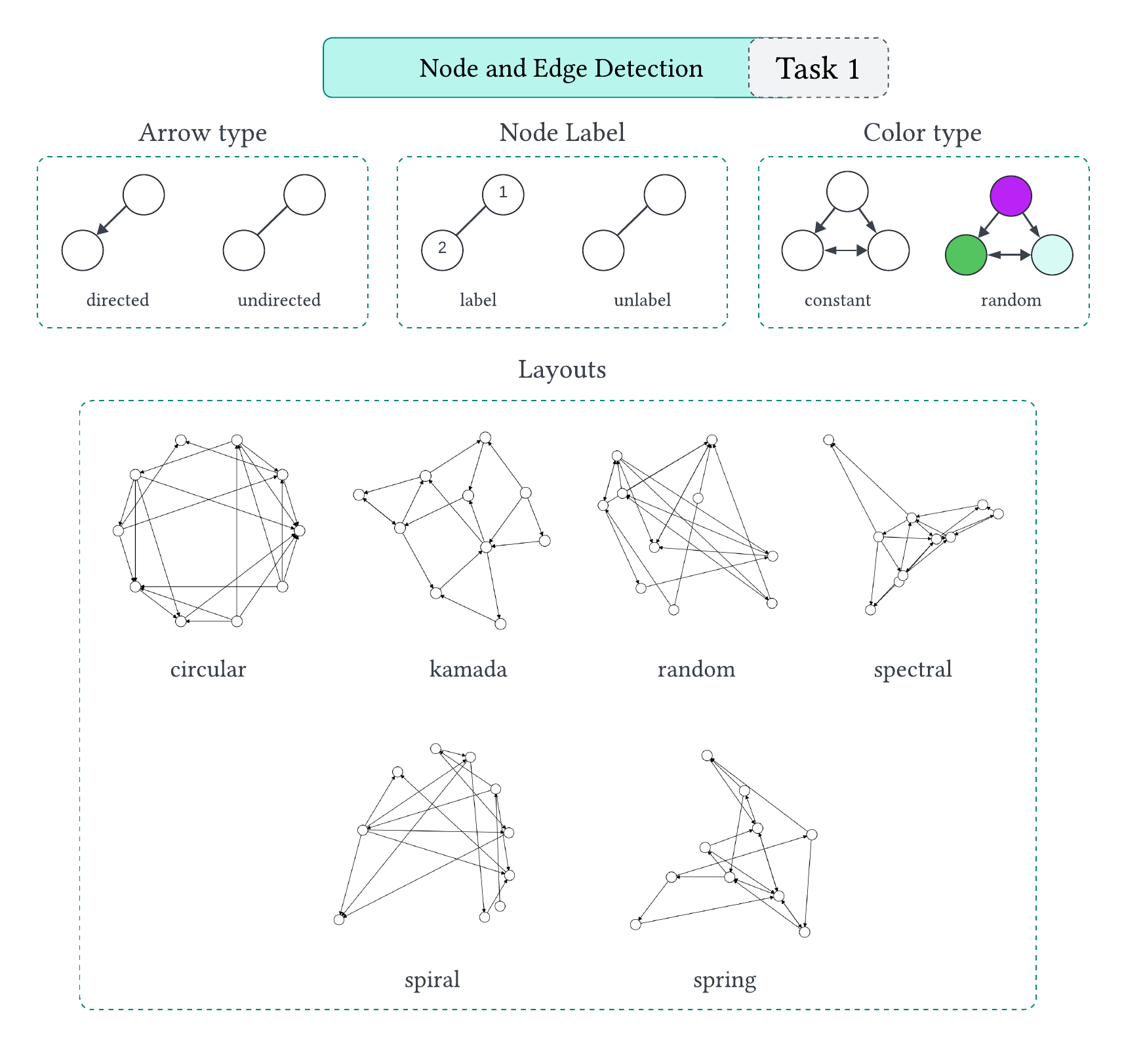}
    \caption{Available configurations for generating graph images to evaluate node and edge detection capabilities.}
    \label{fig:task1}
\end{figure}

A proficient LVLM for graph analysis should be able to accurately detect and count existing elements before progressing to more complex reasoning tasks. This basic capability is a prerequisite for more sophisticated analyses. For this reason, we have incorporated a variety of stylistic variations into the generation options for this task, as shown in Figure~\ref{fig:task1}. 

These variations include not only the graph layout but also the use of arrows, labels, and distinct node colors. An effective LVLM should be able to follow the instructions provided by means of the prompt. For instance, when counting elements, attributes such as node labels, directed or undirected edges, or varying colors should not influence the count. However, as we will explore in the next section, not all LVLMs meet these criteria.

Visual benchmarks often assume that visual representations are flawless~\cite{wei2024gitagraphvisualtextual, li2024visiongraphleveraginglargemultimodal}. In reality, visual imperfections are common, especially as networks grow larger, and can occur either intentionally or unintentionally. Creating visualizations may be error-prone. Since humans naturally adapt to and work around these visual defects, \visgraphvar\ deliberately incorporates and sometimes introduces such imperfections into the generated graph images.

\begin{figure}[!t]
    \centering
    \includegraphics[width=0.7\linewidth]{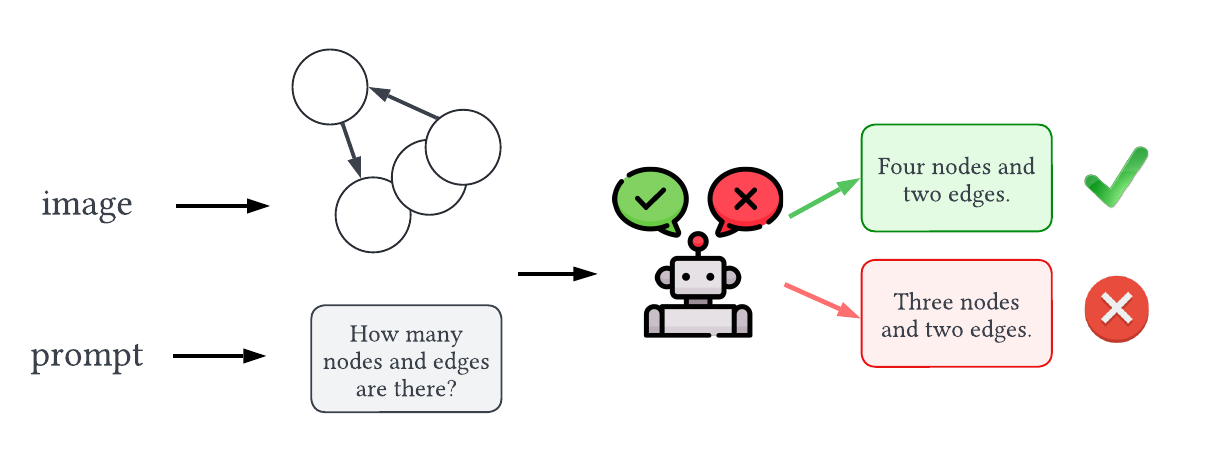}
     \caption{LVLM execution of Task 1 with overlapping nodes and prompt input.}
    \label{fig:task1-overlap}
\end{figure}

Node and/or edge overlap is a prime example of these real-world visual challenges. In our evaluation approach, we expect LVLMs to provide counts that closely approximate the actual number of elements, even when some are overlapping. For instance, if an image contains 10 nodes with three overlapping ones, an LVLM that detects 9 nodes demonstrates better performance than one that only identifies 8. This mirrors human visual processing---we can typically make reasonable estimates of overlapping elements, even when faced with some uncertainty. We expect LVLMs to match or exceed this human capability, particularly in cases of slight overlap (see Figure~\ref{fig:task1-overlap}). Because these visual imperfections serve as a critical test that clearly reveals a visual model's true robustness.

\subsubsection{Task 2: Classification}
\label{subsec:task2}

In addition to mastering the task of detecting and counting nodes and edges, LVLMs should be able to classify a graph in the sense of determining its type. This task presents greater complexity as it requires global image analysis, considering edge positions and directions. Such capabilities are crucial for comprehensive graph analysis. \\

%recognize broader patterns and structures formed by these elements. 

This classification task is essential, as shown already in related applications. LVLMs have been used, for example, to interpret and analyze various types of charts~\cite{masry2023unichart, islam2024largevisionlanguagemodels}, including boxplots, pie charts, etc. This process begins with classifying the detected chart type, a crucial step before conducting any detailed analysis. Furthermore, classification is essential in real-world applications, such as the classification of medical images~\cite{hu2024omnimedvqanewlargescalecomprehensive}. Thus, classification---whether for visual graphs, charts, or real-world applications---is a fundamental task for any foundational vision model.

\begin{figure}[!t]
    \centering
    \includegraphics[width=0.8\linewidth]{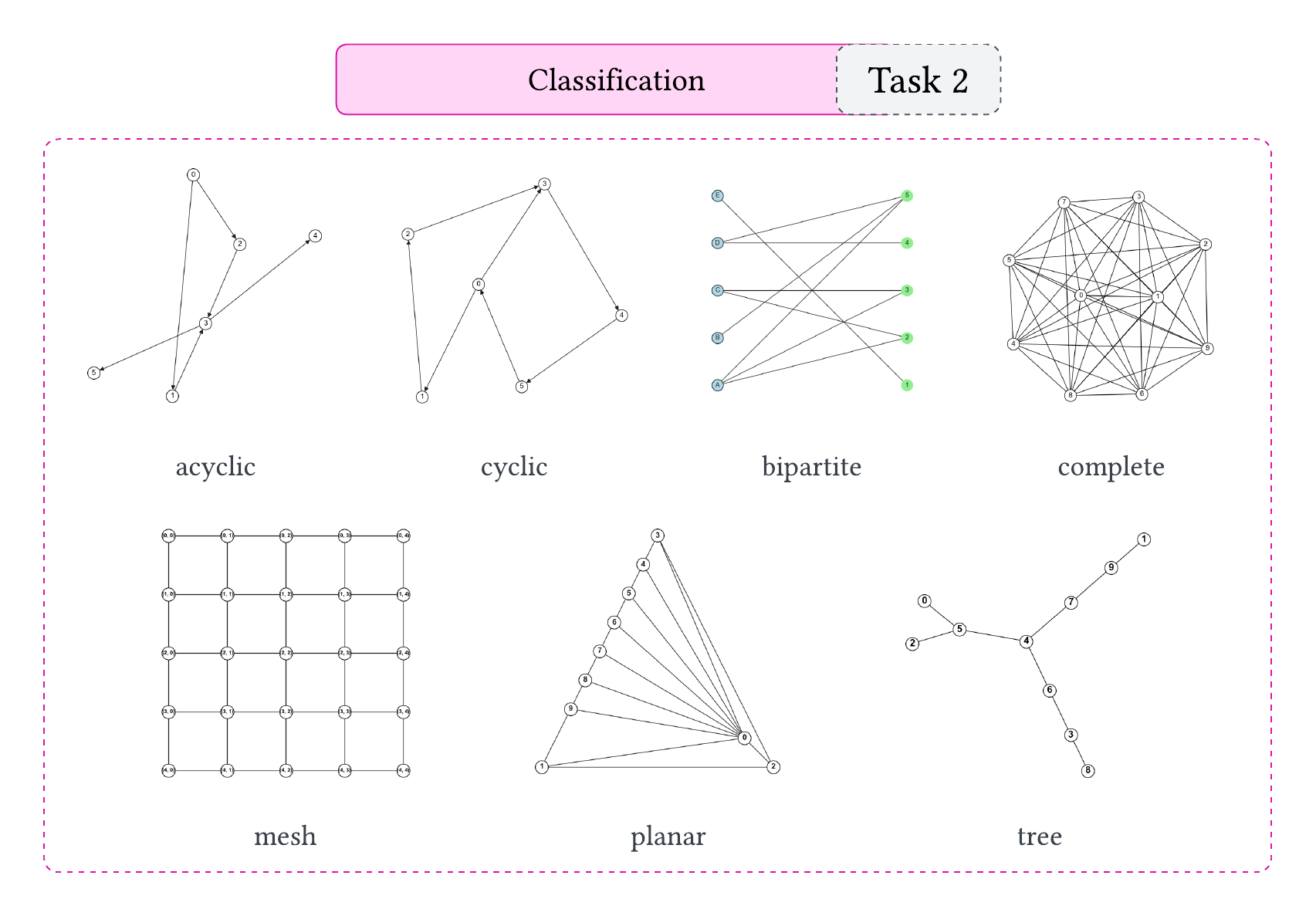}
     \caption{Seven different types of graphs.}
    \label{fig:task2}
\end{figure}

To evaluate an LVLM's proficiency in understanding these higher-order relationships between nodes and edges, we examine its ability to classify seven fundamental graph types (see Figure~\ref{fig:task2}):

\begin{enumerate}[topsep=5pt]
    \item \textbf{Acyclic graphs:} graphs without loops, that is, without any path starting from a node and returning to that node.
    \item \textbf{Cyclic graphs:} graphs containing one or more loops.
    \item \textbf{Bipartite graphs:} the set of nodes is partitioned into two distinct node sets with edges only between these two sets.
    \item \textbf{Complete graphs:} every node connects to all others.
    \item \textbf{Meshs:} regular grid-like structures.
    \item \textbf{Planar graphs:} graphs that can be drawn without edge crossings.
    \item \textbf{Tree:} Hierarchical graphs branching from a root.

\end{enumerate}

Each graph type presents unique analytical challenges. The selection included graphs that are clearly distinct from each other, such as trees and complete graphs, as well as graphs that are similar but with subtle differences, such as cyclic and acyclic graphs. This variety augments the thoroughness of the LVLM's analysis. For example, identifying cycles requires the model to trace edge paths and detect closed loops, while recognizing bipartite structures demands understanding node groupings and their interconnection patterns. Similarly, tree classification requires comprehending hierarchical relationships and directional flow from root to leaves. A LVLM should possess these analytical capabilities to accurately classify graph structures and understand their inherent properties. This understanding forms a crucial foundation for more complex visual reasoning tasks.

The VisionGraph benchmark~\cite{li2024visiongraphleveraginglargemultimodal}, for example, includes a classification task within its reasoning tasks, specifically to detect the presence of a cycle in a graph. In contrast, our benchmark generator offers six additional types of graphs for classification. This difference arises because VisionGraph emphasizes reasoning tasks, whereas \visgraphvar\ reserves reasoning tasks primarily for Task 7; see Section~\ref{reasoning}.

\subsubsection{Task 3: Segmentation}

Beyond the classification of graphs, LVLMs must identify critical regions within graph images, such as cut-edges (or bridges). A cut-edge is an edge whose removal increases the graph's number of connected components---effectively segmenting the graph into two or more disconnected subgraphs~\cite{Grinberg2023-wj, Benjamin2015-kj}. In the introduction to their article, \citet{articleCamilusVK} stated the following:
\begin{quote}
Image segmentation can simply result in an image partition composed of relevant regions.    
\end{quote}
The task of dividing a graph into relevant regions (or segments) has numerous practical applications, including analyzing graph connectivity~\cite{4587440}, optimizing routing~\cite{Miller2022-bs}, and identifying potential failure points and dependencies~\cite{10.5555/982792.982803}.

\begin{figure}[!t]
    \centering
    \includegraphics[width=0.8\linewidth]{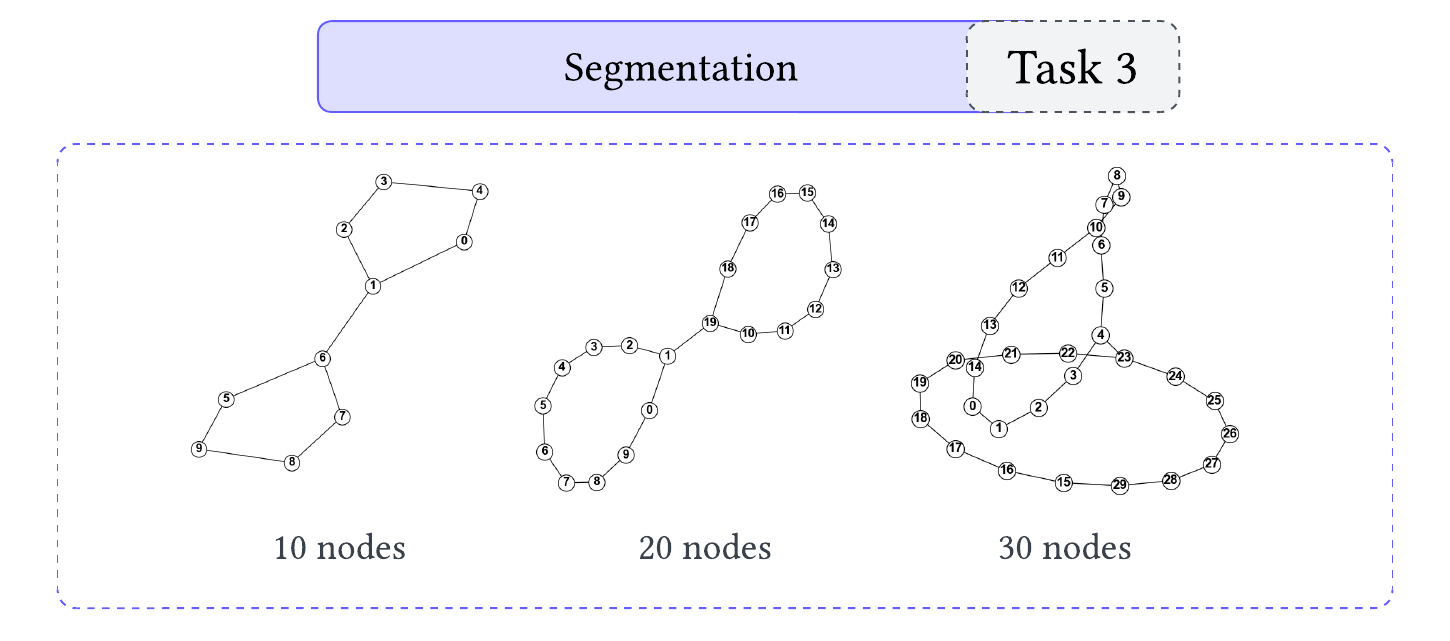}
     \caption{Networks with an increasing number of nodes and a single cut-edge: the graph on the left has cut-edge (6, 7); the one in the center has (1, 19); and the most complex one to detect, on the right, has (4, 23).}
    \label{fig:task3}
\end{figure}

One of the main challenges is accurately detecting cut-edges (bridges). As shown in Figure~\ref{fig:task3}, when the number of nodes increases, finding the cut-edge becomes even more difficult due to inevitable element overlaps. This difficulty is present even for the human eye. Therefore, having a quick visual detection method that identifies which nodes form a bridge would greatly aid in the analysis. 

An LVLM capable of correctly detecting a bridge suggests that it can analyze the graph not only at a global level (considering all connections between nodes) but also identify the specific two nodes whose removal would increase the number of connected components in the graph. This capability demonstrates a nuanced understanding of both the graph’s structure and critical points that impact connectivity.

\subsubsection{Task 4: Pattern Recognition}

\begin{figure}[!t]
    \centering
    \includegraphics[width=0.8\linewidth]{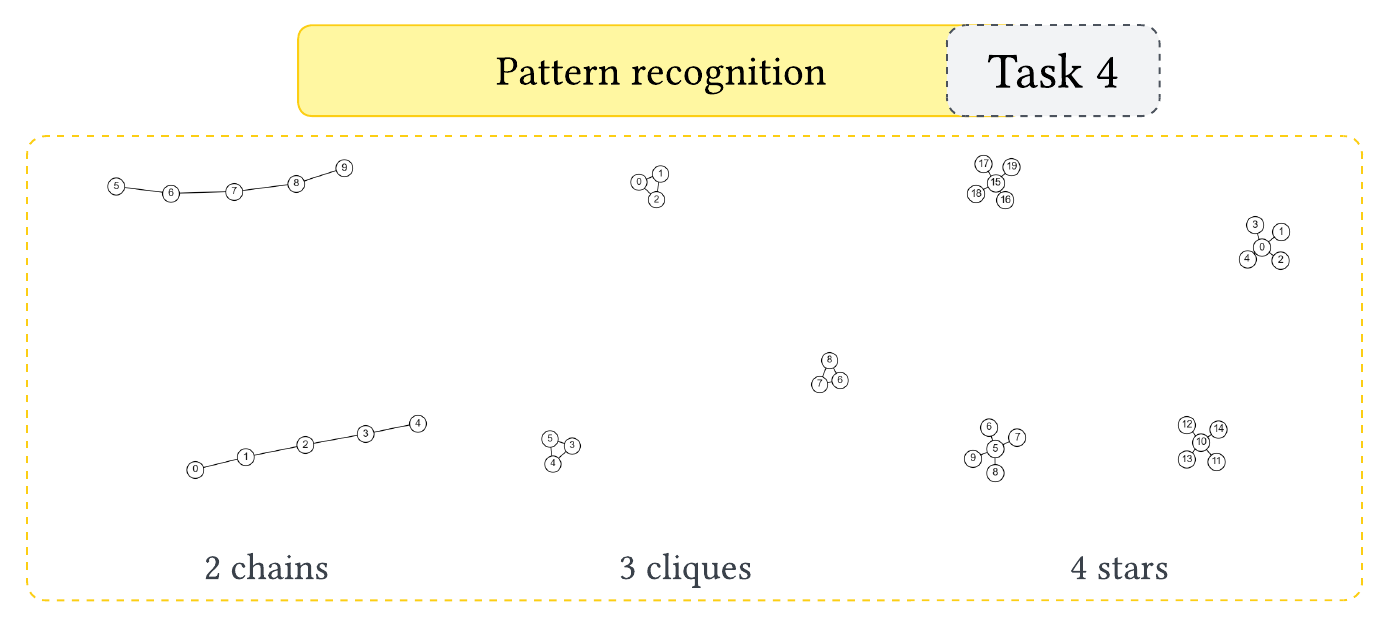}
     \caption{Three graphs with different types of patterns.}
    \label{fig:task4}
\end{figure}

Identifying a cut-edge requires precise, targeted global analysis, but recognizing and counting specific patterns within a graph introduces an even higher level of complexity. Task~4 involves more than a surface-level scan; the LVLM must recognize each unique pattern formed by nodes and edges, classify them accurately, and then provide a count (see Figure~\ref{fig:task4}). This step builds upon Task~2 (classification, Section~\ref{subsec:task2}) by requiring an additional depth of pattern recognition. Such analysis is especially useful for disconnected graphs, where understanding the isolated cluster structures is crucial.

Recognizing patterns in graphs given as mathematical objects (rather than in the form of images) is already a computationally demanding task~\cite{Shi2020-pb}. LLVMs could help reduce computation time if these patterns could be identified through image analysis.

This task also aims to test the memory retention capabilities of LVLMs, a growing area of study due to the integration of memory mechanisms in LLM agents \cite{zhang2024surveymemorymechanismlarge}. Accurately counting recognized patterns requires the model to store previously detected structures and maintain a tally, which is essential for consistency. A less powerful LVLM might identify only a single pattern within the entire graph or detect multiple patterns without distinguishing between types. Thus, this task tests two interdependent levels: the initial recognition of patterns and the subsequent tracking of their occurrences.

\subsubsection{Task 5: Link Prediction}

Apart from memory retention and pattern recognition, there are other crucial tasks in graph analysis. A vital aspect concerns reasoning, with the example of link/edge prediction. When examining a graph, its structure and the positioning of nodes and edges (determined by the layout) may potentially suggest missing connections. By analyzing the overall topology, we can identify nodes with similar structural patterns that could logically be connected. This concept, known as link prediction, emerges when nodes sharing similar topological characteristics are not yet connected but show potential for connection~\cite{mao2024revisitinglinkpredictiondata}.

\begin{figure}[!t]
    \centering
    \includegraphics[width=0.8\linewidth]{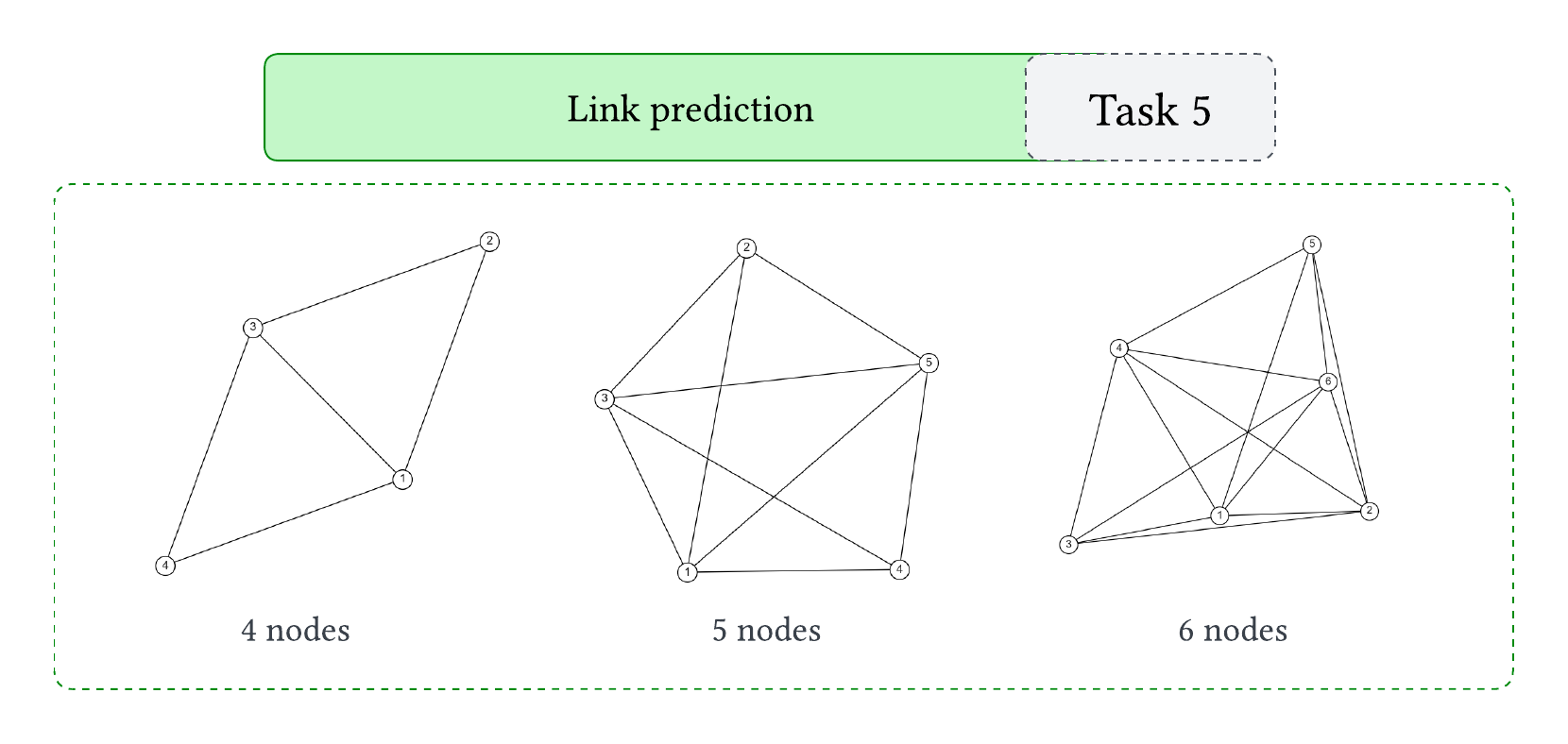}
     \caption{Three types of graphs with different numbers of nodes for which the LVLM is expected to predict a missing link/edge. The missing link on the left is (4,2), the one in the center is (2, 4), and the most complex case is (3, 5), on the right.}
    \label{fig:task5}
\end{figure}

Interestingly, the complexity of link prediction does not necessarily increase with the number of nodes (see Figure~\ref{fig:task5}). Smaller graphs can sometimes present greater challenges for prediction. This counterintuitive situation occurs when the model struggles to identify replicable patterns due to fewer examples of existing connections. The sparsity of edges in smaller graphs can make it harder to establish reliable prediction patterns.

Link prediction differs from other graph analysis tasks as it requires a deeper level of reasoning from LVLMs. Rather than simply recognizing existing patterns or memorizing structures, the model must perform the following steps:

\begin{itemize}[topsep=5pt]
    \item Analyze the global graph structure
    \item Identify similar topological patterns across different node pairs
    \item Make logical inferences about potential connections
    \item Consider both local and global network characteristics
    \item Apply structural similarity principles to predict missing links
\end{itemize}

Currently, numerous research efforts focus on graph link prediction, though these approaches typically work with graphs as mathematical objects rather than with visual representations~\cite{zhang2018linkpredictionbasedgraph,li2023evaluatinggraphneuralnetworks}. The advanced reasoning requirement makes link prediction a particularly valuable benchmark for evaluating an LVLM's capability to understand and reason about complex network structures.

\subsubsection{Task 6: Reasoning}\label{reasoning}

If link prediction already requires the model to perform a certain level of reasoning, asking it to apply an algorithm to the graph displayed in the image significantly increases the difficulty. This is no longer just a visual task that can be inferred from the position of nodes and edges. An example is the problem of finding shortest paths, where the model must identify nodes, detect edge directions, store edge weights in memory, and then determine the shortest path between a source and destination. This task will test both the reasoning and memorization capabilities of the LVLMs.

Moreover, this type of task requires applying numerical analysis to visual information. Generally, LLMs struggle with reasoning because their inferences are purely statistical and based on their training data~\cite{huang2023reasoninglargelanguagemodels}. This challenge is very complex for LVLMs because, in addition to processing text input, the model must reason about an image containing many visual elements, each of which can play a different role and change the ``direction'' of the reasoning process. In contrast to LLMs that generally only receive textual data to make inferences, an LVLM has to simultaneously process and reason about visual elements in an image, understand how they relate to each other, and integrate this understanding with any textual input---making the reasoning task substantially more complex. However, new prompting techniques like self-reflection~\cite{sahoo2024systematicsurveypromptengineering}, while not completely solving the problem, do improve results in reasoning tasks~\cite{madaan2023selfrefineiterativerefinementselffeedback}. These techniques have already been applied to LVLMs~\cite{cheng2024visionlanguagemodelsselfimprovereasoning}. 

\begin{figure}[!t]
    \centering
    \includegraphics[width=0.8\linewidth]{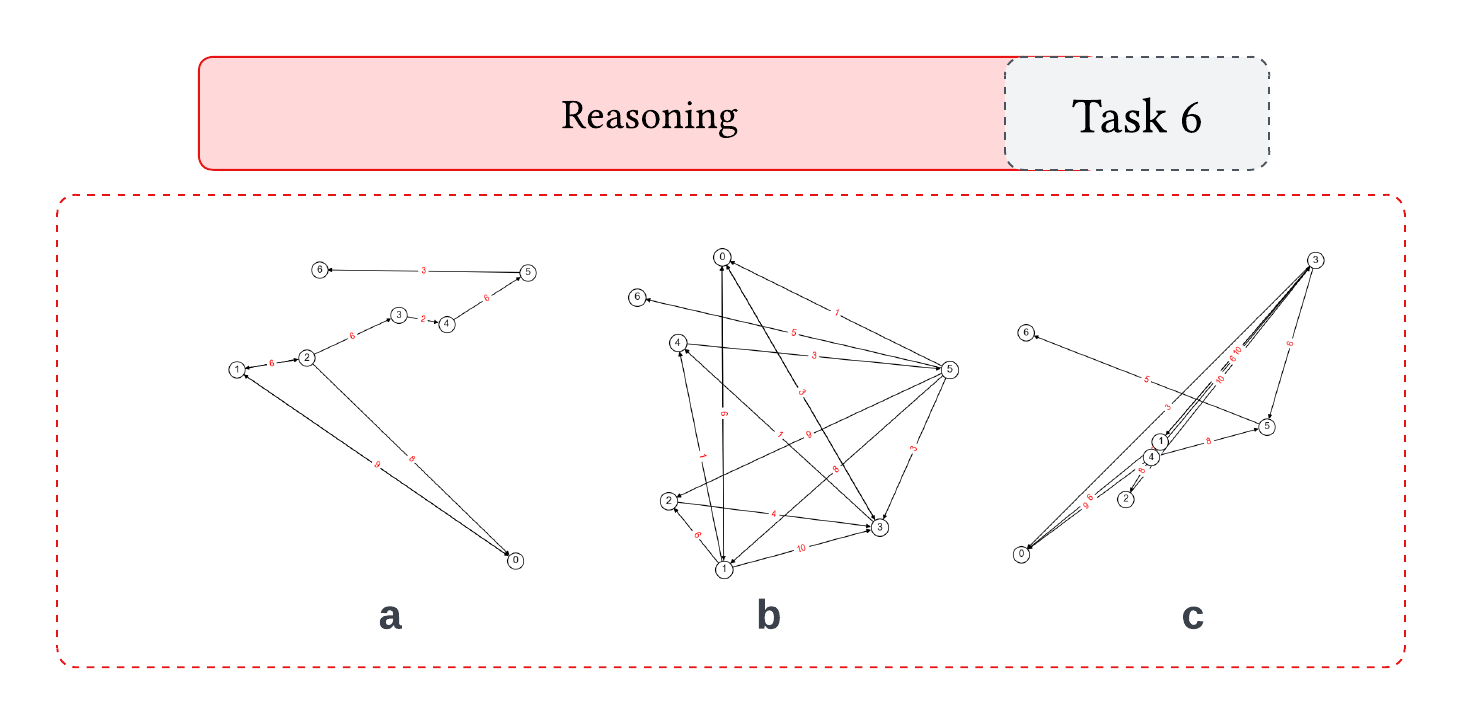}
     \caption{Three graphs with varying levels of interpretive difficulty in identifying shortest paths. (a) and (b) are simpler due to the lack of overlap between nodes and edges, whereas (c) makes it very hard to locate each node along a shortest path due to element overlap.}
    \label{fig:task6}
\end{figure}

Examples of the shortest path-finding task are shown in Figure~\ref{fig:task6}. In particular, graphs created by \visgraphvar\ with weighted edges and labeled nodes are shown. Given a source-target node pair, the LVLM is asked to determine the shortest path regarding the sequence of nodes. Cases (a) and (b) are expected to be easier to solve since we humans can easily detect the shortest path, although it is somewhat more difficult in case (b) compared to (a). Case (c) has overlapping nodes, edges, and weights. Therefore, it is very probable that the LVLM will make mistakes here, as even humans might provide the wrong answer. However, this case is useful as it assesses how the LVLM behaves in such extreme cases. Moreover, this task shows how visual complexity affects both human and LVLM performance in graph analysis tasks, using specific examples to demonstrate increasing levels of difficulty---from clearly visible paths to overlapping elements where accurate analysis becomes very hard. 

The VisionGraph~\cite{li2024visiongraphleveraginglargemultimodal} benchmark includes more reasoning tasks than \visgraphvar, including connectivity, topological sorting, maximum flow, bipartite graphs, Hamiltonian paths, and GNN tasks. However, as our results demonstrate (see Section~\ref{experiments}), even the currently best LVLMs ``struggle'' with simpler tasks like shortest path finding. Thus, adding further reasoning tasks may not be fully justified, as there are foundational tasks that models need to handle before advancing to reasoning. %\visgraphvar\ differs in this regard, covering tasks that span before, during, and after visual algorithmic reasoning, providing a more comprehensive scope.
 
\subsubsection{Task 7: Matching}

Determining whether or not two given graphs match is an important task in graph theory~\cite{Candelier2024-hb}. For this purpose, the LVLM must identify the topology of the two graphs and align elements---nodes and edges---to map relationships accurately. This process has applications in a range of fields, including neuroscience~\cite{10.1162/netn_a_00133}, computational biology~\cite{Devkota2023.03.13.532445}, computer vision~\cite{9263681}, and machine learning~\cite{li2019graphmatchingnetworkslearning}.

\begin{figure}[!t]
    \centering
    \includegraphics[width=0.8\linewidth]{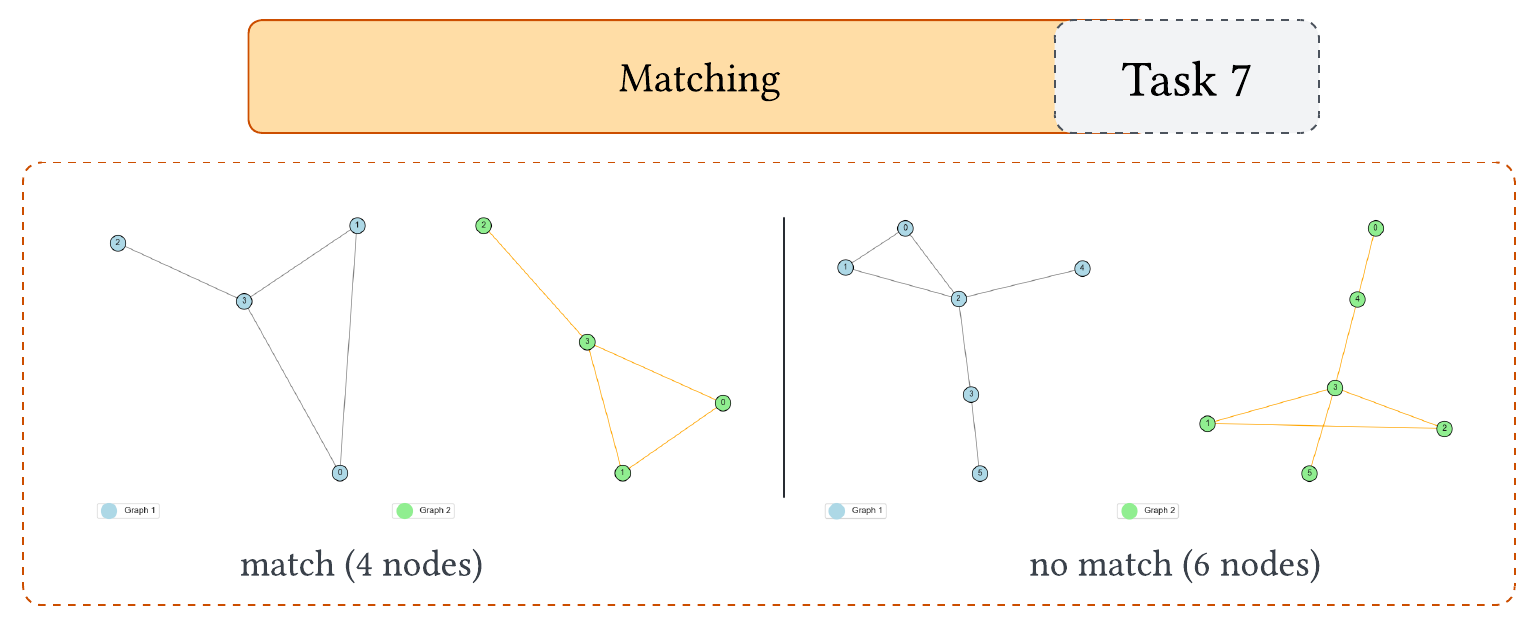}
     \caption{Graph pairs are shown with the goal for the LVLM to identify matches on the left and distinctions on the right. Note that, in this paper, two graphs in the same image are said to match if their structure (including node labels) is equal; that is, only their display style might differ. For this reason, the two graphs on the right do not match, even though they are isomorphic.}
    \label{fig:task7}
\end{figure}

Unlike the previous graph tasks, matching involves analyzing two graphs simultaneously to determine their structural similarity. When nodes with the same labels share the same connection patterns, the graphs are said to match in this paper. Otherwise, they do not match, even though they might still be isomorphic, for example; see the example on the right-hand side of Figure~\ref{fig:task7}. The core verification steps for graph matching by an LVLM should include:

\begin{itemize}[topsep=5pt]
    \item Analyzing the overall structure of both graphs
    \item Identifying and comparing the number of nodes and edges
    \item Verifying if connected nodes are equivalent (by checking their labels)
    \item Taking into account edge directionality
    \item Omitting stylistic details such as node colors, label size, and the layout that determines node positioning
\end{itemize}

In this task, as shown in Figure~\ref{fig:task7}, an LVLM should ideally filter out visual details that do not impact the graph's topology---such as colors, node sizes, or layout. When processing the graph image, the LVLM must focus on the prompt, which clarifies that the task centers on matching, guiding the model to prioritize relevant structural information.

\subsection{Dataset Configuration and Statistics}\label{dataset-stat}

Next, \visgraphvar\ was used to generate a diverse dataset of graph images whose detailed statistical breakdown is shown in Table~\ref{tab:dataset_distribution}. This dataset comprises a total of 990 images, each measuring 600x600 pixels. The image count varies through tasks and depends on the settings chosen in \texttt{VisGraphVar}, as detailed below.

\begin{table}[!t]
    \centering
    \caption{Dataset distribution by task, with image counts and percentages.}
    \label{tab:dataset_distribution}
    \begin{tabular}{l|l|c} \toprule
        \textbf{Task number} & \textbf{Task} & \textbf{Images} \\
        \midrule
        1 & Detection            & 560 (56.57\%) \\
        2 & Classification       & 70  (7.07\%)  \\
        3 & Segmentation         & 30  (3.03\%)  \\
        4 & Pattern Recognition  & 210 (21.21\%) \\
        5 & Link Prediction      & 30  (3.03\%)  \\
        6 & Reasoning            & 30  (3.03\%)  \\
        7 & Matching             & 60  (6.06\%)  \\
        \bottomrule
    \end{tabular}
\end{table}

\begin{itemize}[topsep=5pt]
    \item Task 1 (Detection) includes 560 images, considerably more than other tasks. This is due to enhanced stylistic diversity (node colors, arrow types, layouts, and node labels). All graphs in these images contain 10 nodes, and the number of edges is determined by the probability percentage (2\%) of adding an edge between any pair of nodes. Remember, that both parameters are configurable in \visgraphvar.
    \item Task 2 (Classification) consists of 70 images, representing seven graph types (tree, planar, mesh, cyclic, complete bipartite, and acyclic). The number of nodes and edges depends on the type of graph. In our dataset, the number of nodes does not exceed 10, and the number of edges varies based on the type of graph.
    \item Task 3 (Segmentation) contains 30 images with varying node counts (10, 20, and 30). The number of edges is calculated based on node count, creating a graph where two subgraphs connect via a single cut-edge.
    \item Task 4 (Pattern Recognition) has 210 images with three pattern types (chain, clique, and star) in varying quantities (2, 3, and 4 patterns per image). The number of edges is automatically calculated based on node count and desired pattern.
    \item Task 5 (Link Prediction) includes 30 images, each with a graph containing a unique node count (4, 5, and 6 nodes). Complete graphs are created, but with one edge removed (the one to be predicted).
    \item Task 6 (Reasoning) has 30 images, featuring graphs with 5, 6, or 7 nodes. By default, a random layout is used. Similar to Task 1, the number of edges is determined by a probability percentage (3\%) for adding an edge between any pair of nodes. The edge weights are random values between 1 and 10, which is also configurable in \visgraphvar.
    \item Task 7 (Matching) includes 60 images, showing pairs of graphs that match and others that do not match. The graphs in these images have 4, 5, or 6 nodes. Like in Tasks 1 and 6, edges are added based on a 4\% probability of being included between any pair of nodes. 
\end{itemize}

For each task, 10 images were generated for each variation regarding the graph's visual style. This allows for an average calculation of each metric (see Section~\ref{metrics}). For example, in Tasks 3, 5, and 6, where graphs are of three different sizes regarding the number of nodes, there are a total of 30 images, corresponding to the three unique visual variations.

As managing Task 1 well is a prerequisite for all other tasks, we decided to test various visual styles in this context, creating a dataset designed to pose a challenge for any LVLM.

\subsection{Metrics}
\label{metrics} 

As described in the previous section, each visual combination is evaluated using a set of 10 images. We employ three different metrics, which differ by task, so not all tasks share the same evaluation criteria. Each metric is normalized on a scale from 0 to 1, where 1 indicates optimal performance (complete alignment with the ground truth), and 0 indicates the lowest performance (no alignment with the ground truth). Further details are provided below. 

\paragraph{Mean absolute error (MEA).} In Task 1, we used the MAE because we are interested in knowing the degree of error in predicting the number of nodes and edges that the model infers from the image.

\begin{equation}
\begin{aligned}
\text{MAE} &= \frac{1}{n}\sum_{i=1}^{n}|y_i - \hat{y}_i| \\
\text{NMAE} &= 1 - \min\left(\frac{\text{MAE}}{\text{Range}}, 1\right) \quad, \text{where} \\ 
\text{Range} &= 
    \begin{cases}
        \max(y) - \min(y) & \text{if } \max(y_i) \neq \min(y_i)  \\
        1 & \text{if } \max(y_i) = \min(y_i)
    \end{cases}
\end{aligned}
\end{equation}

Hereby, $y_i$ is the actual number of elements (either nodes or edges) that appear in the image, while $\hat{y}_i$ is the value predicted by the model. Moreover, NMAE is the normalization of MAE---using \text{Range}---to ensure that predictions closer to the true values approach a value of 1, while those further away approach 0.

\paragraph{Accuracy.} Following \citet{Zou2024-tt}, we assess model responses in tasks 2-5, and 7 using an accuracy metric, but we extended it to account for both exact and partial matches. The accuracy is calculated as:

\begin{equation}
\begin{aligned}
\text{Accuracy} &= \frac{1}{n}\sum_{i=1}^{n} m_i \quad, \text{where}\\
m_i &=
\begin{cases}
1 & \text{if } y_i = \hat{y}_i \\
0.5 & \text{if } y_i \text{ partially matches } \hat{y}_i \\
0 & \text{otherwise}
\end{cases}
\end{aligned}
\end{equation}

This formulation allows us to account for partial correctness in tasks where a prediction may be partially correct. For instance, in task 5 (link prediction), an accuracy score of 0.5 is assigned if only one of the two nodes of the missing edge is correctly predicted by the model. The same principle applies to other tasks where partial matches are possible.

\paragraph{Jaccard Index.} Task 6 (reasoning) requires the model to find a shortest path regarding a sequence of nodes. Therefore, we apply the Jaccard Index to provide a similarity value between the actual nodes that form a true path with the path predicted by the model.

\begin{equation}
    \begin{aligned}
    \text{Jaccard Index} &= \frac{|P_\text{true} \cap P_\text{pred}|}{|P_\text{true} \cup P_\text{pred}|} \quad,
    \end{aligned}
\end{equation}

where $P_\text{true}$ represents the set of nodes in the true path and $P_\text{pred}$ represents the set of nodes in the predicted path. That way, even if a model only correctly predicted four nodes of a five-node path, it would have a higher score than a model that only correctly predicted two nodes out of a five-node path.

\subsection{Prompt design}

Like all multimodal models, LVLMs require both an image and a text prompt to operate. We will be testing two different prompting approaches: (1) zero-shot, where the model makes direct predictions without examples)~\cite{kojima2023largelanguagemodelszeroshot}; and (2) chain-of-thought, where the model explains its reasoning step by step~\cite{wei2023chainofthoughtpromptingelicitsreasoning}. \citet{vatsal2024surveypromptengineeringmethods} identified up to 39 prompting strategies available for natural language processing tasks; however, many of these are derivatives of the zero-shot and chain-of-thought approaches, which justifies our selection. Thus, this approach requires two distinct prompts for each task, resulting in 14 prompts in total. To streamline this process, we developed a single basic prompt per task and then used an LLM to create two versions of each---one using the zero-shot format and another using the chain-of-thought format. We use these two prompt versions alongside each image to test all considered LVLMs. Moreover, we request all model responses in JSON format (as specified in the prompts) to facilitate a later analysis. The original generic prompts (defined by hand) and the LLM-generated versions used for dataset evaluation are included in Appendix~\ref{app:prompt-design}.

\section{Experiments and Evaluation}
\label{experiments}

In this section, we evaluate six state-of-the-art LVLMs across seven tasks using the dataset described before and generated by \visgraphvar. More specifically, Section~\ref{experimental-methodology} details our experimental setup, execution environment, and the rationale behind our LVLM selection. Finally, we present our quantitative analysis in Section~\ref{results}, followed by qualitative observations in Section~\ref{observations}.

\subsection{Environment Setup and LVLM Configuration}
\label{experimental-methodology} 

For the evaluation, we used the 990 images from our dataset categorized across seven tasks (see Section~\ref{dataset-stat}). Model evaluation was conducted through the OpenRouter API\footnote{\url{https://openrouter.ai}.}, which facilitates efficient multi-model execution. We selected LVLMs based on their performance ranking in the Chatbot Arena LLM Leaderboard (vision)\footnote{\url{https://lmarena.ai}.}, specifically: GPT-4o-2024-08-06~\cite{wu2024gpt4ovisualperceptionperformance}, Gemini-Pro-1.5~\cite{geminiteam2024gemini15unlockingmultimodal}, Claude-3.5-Sonnet~\cite{anthropicIntroducingClaude}, Llama-3.2-90B-Vision-Instruct~\cite{dubey2024llama3herdmodels}, Qwen-2-VL-72B-Instruct~\cite{yang2024qwen2technicalreport}, and Pixtral-12B~\cite{agrawal2024pixtral12b} as of October 2024. The Chatbot Arena, developed by researchers at UC Berkeley SkyLab and LMSYS, is an open-source platform for AI evaluation through human preference data. Its LVLM evaluation framework has garnered substantial user engagement, with over \num{130000} votes across 38 different LVLMs. As mentioned above, for each image, we employed two prompting strategies: zero-shot (0-shot) and chain-of-thought (CoT). This resulted in \num{1980} evaluations per model (990 images per 2 prompting strategies), totaling \num{11880} evaluations across all six models. For all models, we used OpenRouter's default parameter settings (e.g., temperature) without any modifications or tuning.

\subsection{Results}
\label{results}

In this section, we present a detailed comparative analysis of the obtained results. These results are shown regarding the utilized metrics (see Section~\ref{metrics}). However, note that, for the sake of a better understanding, metric scores are shown in terms of percentages. ~\Cref{fig:average-all-model,fig:claude-3.5-sonnet-best} provide summarized results to easily identify the leading models across all tasks.  Additionally, Figure~\ref{fig:heatmap-comparison} shows the impact of prompt strategies on the results.

\subsubsection{Task-Specific Performance Analysis} 

\begin{figure}[!t]
    \centering
    \includegraphics[width=1\linewidth]{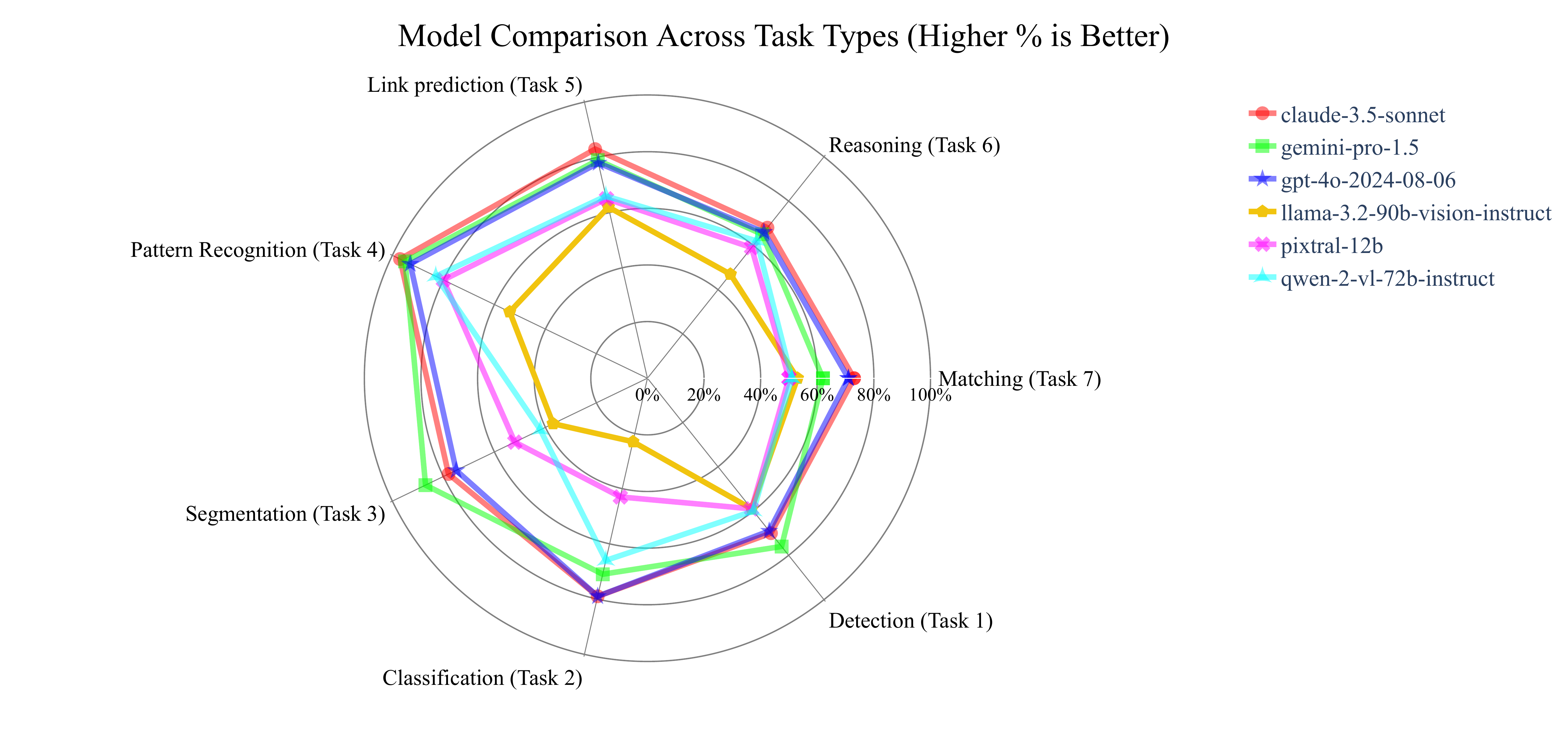}
    \caption{An overview of LVLM performance across the seven tasks (complete dataset).}
    \label{fig:overview-all-task-models}
\end{figure}

In Figure~\ref{fig:overview-all-task-models}, we observe that Claude-3.5-Sonnet, Gemini-Pro-1.5, and GPT-4o perform similarly across most tasks (70\%-80\% overall accuracy rate). However, significant differences are observed in Task 7 and Task 2, where Claude-3.5-Sonnet and GPT-4o outperform Gemini-Pro-1.5. In contrast, Gemini-Pro-1.5 and GPT-4o exhibit a slightly lower performance in Task 1 and Task 3 when compared to Claude-3.5-Sonnet. Notably, Gemini-Pro-1.5 clearly outperforms all other models in Task 3 and shows a slight advantage in Task 1. These results highlight the general advantage of proprietary models over open-weight alternatives.

An interesting observation is that all models show a rather high performance on Task 4 when compared to their performance on other tasks. Moreover, another eye-catching finding is Qwen-2-vl-72B's performance in Task 6, where it nearly matches the top three models. Another significant observation concerns Llama3.2-90B, which, despite having substantially more parameters than Qwen-2-VL-72B and especially Pixtral-12B, exhibits markedly lower performance than both models and ranks below all other tested models. This outcome confirms the suspicion that simply increasing the number of model parameters does not necessarily lead to improved performance in visual analysis tasks; in fact, as noted by~\citet{mckenzie2024inversescalingbiggerisnt}, inverse scaling may occur due to flaws in the training objective and data.

\subsubsection{Performance Distribution Analysis by Task}

\begin{figure}[!t]
    \centering
    \includegraphics[width=0.8\linewidth]{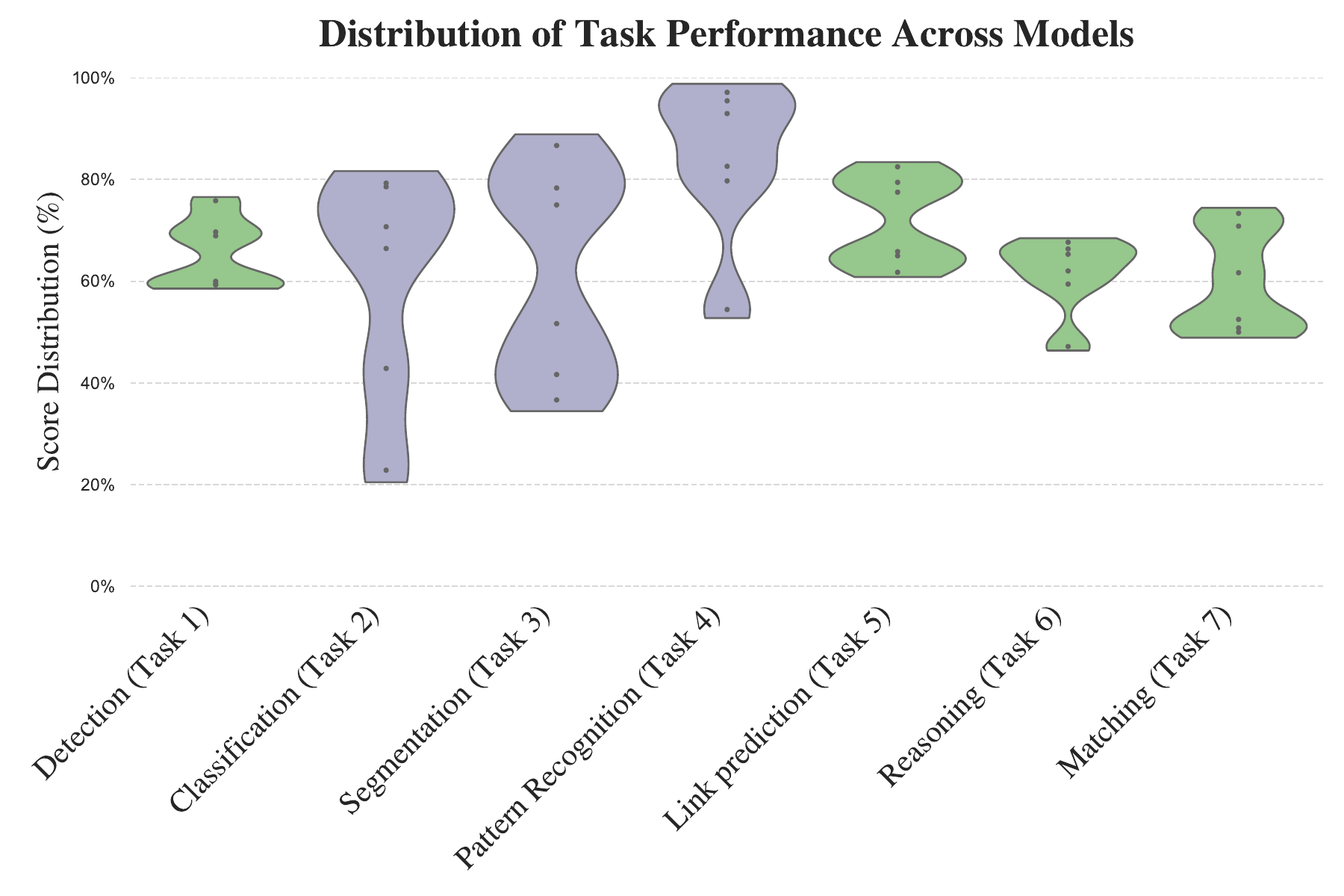}
     \caption{The distribution of average scores across the six LVLMs for each task. The violin plot is configured with $\texttt{bw\_adjust}=0.5$ (which adjusts the bandwidth of the kernel density estimation, making the plot more detailed) and $\texttt{cut}=0$ (which ensures the plot is limited to the range of the data without extending beyond it) using the Seaborn library in Python.}
    \label{fig:distribution-task}
\end{figure}

Figure~\ref{fig:distribution-task} employs violin plots to visualize the performance distribution patterns across our evaluated models for all seven tasks. The visualization combines two key elements: individual points representing each model's average performance score, and varying plot widths that indicate the density of scores at each performance level. This dual representation enables comprehensive analysis of both individual model performance and the overall distribution patterns across different tasks.

The green violin plots for Tasks 1, 5, 6, and 7 exhibit a narrow and condensed shape, with an approximate score distribution height of $\sim20\%$ on the y-axis. This indicates that the six models performed consistently and with a lower variation on these tasks. The concentrated distribution suggests that the models' responses were more homogeneous and closely aligned for these particular tasks. In contrast, purple violin plots for Tasks 2, 3, and 4 are wider and more dispersed, with an approximate score distribution height ranging from $\sim40\%$ to $\sim60\%$ on the y-axis. The increased width and height of these violin plots signify greater performance variability among the six models for these tasks. The heterogeneous distribution implies that the models generated diverse responses and exhibited varying performance levels on Tasks 2, 3, and 4.

Task 4 reveals a notably asymmetric distribution pattern characterized by concentrated performance scores in the upper range, with five LVLMs demonstrating robust performance ($\geq80\%$). However, the presence of a single outlier at $\sim50\%$ introduces significant dispersion into the distribution, creating a clear performance dichotomy among the evaluated models.

These variations demonstrate \visgraphvar's capability to capture diverse model behaviors, attributable to its comprehensive task design that spans multiple aspects of visual analysis.

\subsubsection{Aggregate Performance Evaluation}
\label{subsec:agg-perf-eva}

\begin{figure}[!t]
    \centering
    \includegraphics[width=0.7\linewidth]{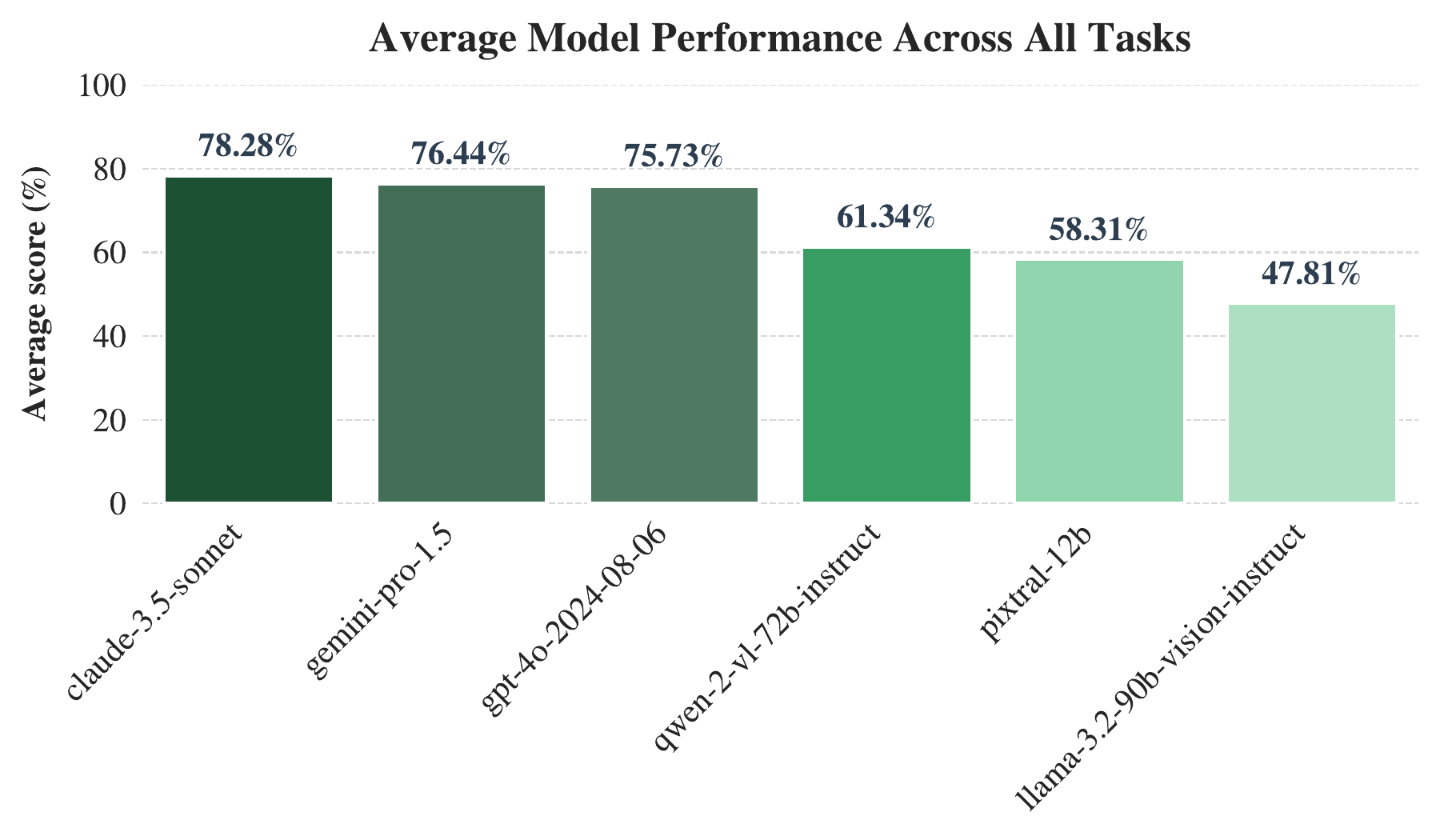}
     \caption{Average LVLM performance (best to worst from left to right) regarding the \visgraphvar\ dataset.}
    \label{fig:average-all-model}
\end{figure}

In Figure~\ref{fig:average-all-model}, it is shown that multimodal models like Claude-3.5-Sonnet, Gemini-Pro-1.5, and GPT-4o exhibit similar performance across all tasks, but with a slight advantage of Claude-3.5-Sonnet. Moreover, there is a significant performance gap of 30.47\% between the top model, Claude-3.5-Sonnet, and the model with the weakest performance, Llama3.2-90B. This trend aligns with the fact that all three top models are proprietary models, further confirming that, at present, these models outperform open-weight models in visual tasks. This confirms the findings of~\cite{gudibande2023falsepromiseimitatingproprietary}, which demonstrated that there is currently no way for an open-weight model to match the performance of a proprietary model without improvements to its underlying base language model.

\begin{comment}
\begin{figure}[H]
    \centering
    % Subfigura (a)
    \begin{subfigure}{1\linewidth}
        \centering
        \includegraphics[width=\linewidth]{figures/overview-results.pdf}
        \caption{}
        \label{fig:task7a}
    \end{subfigure}
    
    \vspace{0.5cm}

    % Subfigura (b)
    \begin{subfigure}{1\linewidth}
        \centering
        \includegraphics[width=\linewidth]{figures/parallel_coordinate_comparison.pdf}
        \caption{}
        \label{fig:task7b}
    \end{subfigure}
    
    \caption{}
    \label{fig:grouped_figures}
\end{figure}
\end{comment}

\begin{figure}[!t]
    \centering
    \includegraphics[width=0.7\linewidth]{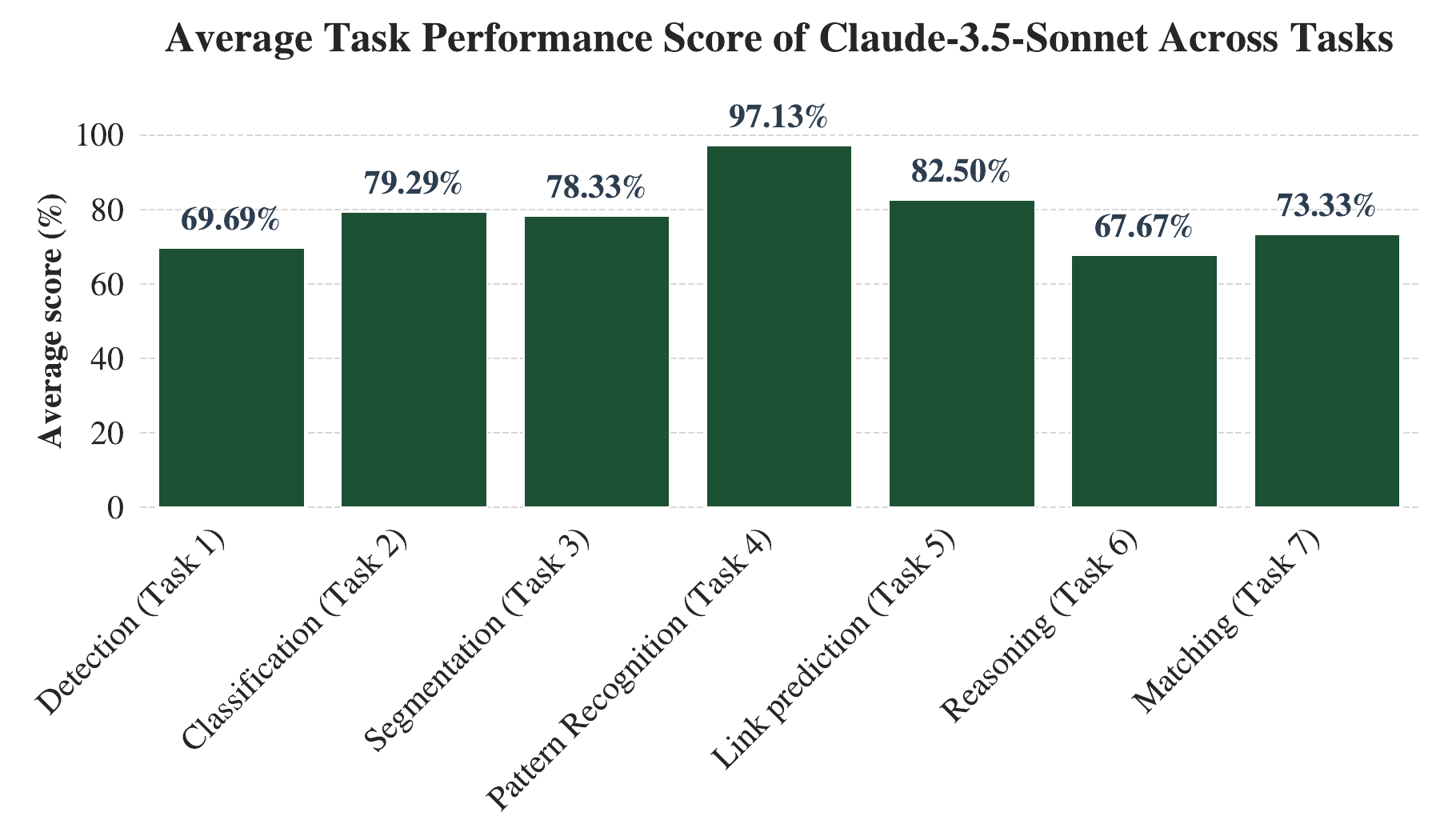}
     \caption{Average performance of Claude-3.5-Sonnet for each task from the \visgraphvar\ dataset.}
    \label{fig:claude-3.5-sonnet-best}
\end{figure}

Figure~\ref{fig:claude-3.5-sonnet-best} shows that Claude-3.5-Sonnet exhibits a mixed performance across tasks. It excels in Task 4 but shows a relatively lower performance in Tasks 1 and 6. Only for Tasks 4 and 5 an average accuracy of over 80\% performance is obtained. These results indicate that, except for Task 4, future LVLMs have significant room for improvement, in particular for what concerns detection (Task 1), Reasoning (Task 6), and Matching (Task 7).

\subsubsection{Prompting Strategy Impact Analysis} 

\begin{figure}[!thp]
    \centering
    \includegraphics[width=0.99\linewidth]{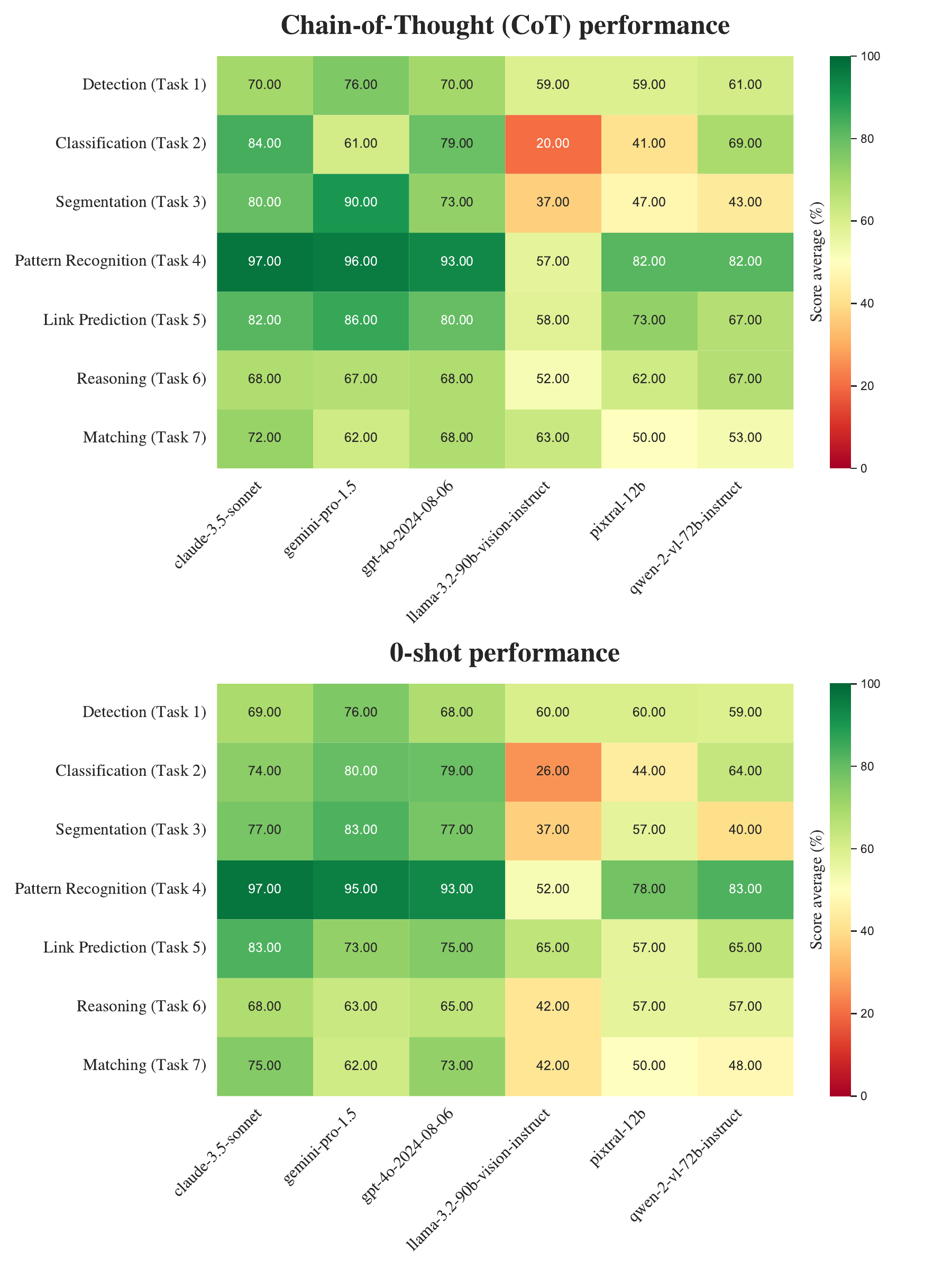}
         \caption{Average scores for each task by prompt strategy, Chain-of-Thought (\textbf{top}) and 0-shot (\textbf{bottom}). Green indicates strong results, while red denotes poor results.}
    \label{fig:heatmap-comparison}
\end{figure}

In Figure~\ref{fig:heatmap-comparison}, we observe that different prompting strategies, in general, only slightly affect model performance. Llama3.2-90B benefits from CoT prompting across most tasks, with two notable exceptions: in Task 2, where 0-shot prompting performs better, and in Task 3, where both strategies yield comparable results.

Claude-3.5-Sonnet, Gemini-Pro-1.5, and GPT-4o demonstrate consistent performance across all tasks, showing minimal variation between different prompting strategies. Similarly, Pixtral-12B and Qwen-2-VL-72B generally exhibit little variation between prompting methods. However, notable exceptions emerge in the analysis: Pixtral-12B demonstrates enhanced performance with CoT prompting compared to 0-shot approaches, achieving a significant 16\% improvement in Task 5. Conversely, in Task 3, we observe a 10\% performance advantage when using 0-shot over CoT prompting.

Overall, and somewhat surprisingly, we do not observe any single prompt strategy consistently outperforming the other. Since 0-shot prompts are easier to create than CoT prompts, we recommend beginning with 0-shot testing; if the results are unsatisfactory, then CoT prompts can be attempted. Other benchmarks in the field support these findings regarding prompting strategies. VisionGraph~\cite{li2024visiongraphleveraginglargemultimodal} similarly fails to demonstrate clear superiority of either prompting approach. GITA~\cite{wei2024gitagraphvisualtextual} takes a different approach, focusing solely on 0-shot prompting and fine-tuned LLMs, while omitting CoT evaluation entirely.

Our analysis provides a more detailed and comprehensive examination of this phenomenon, emphasizing that there is no clear superior approach among the most popular prompting strategies, namely 0-shot and CoT. This finding is important to the ongoing discussion about optimal prompting strategies in visual-language tasks.

\subsection{Observations}\label{observations}

This section analyzes three cases that clarify key aspects of LVLM behavior on our dataset created with \visgraphvar. We first examine \Cref{fig:obs-spectral-layout,fig:layout-spectral}, which reveals crucial insights about spectral layout interpretation in Task 1 (Detection). We then explore \Cref{fig:fail-task7}, highlighting distinctive characteristics in Pixtral-12B's performance on Task 7 (Matching). Finally, \Cref{fig:obs-label-task1} demonstrates the influence of node labeling on model performance and accuracy.

\subsubsection{The Striking Case of Spectral Layout}

\begin{figure}[!th]
    \centering
    \includegraphics[width=0.8\linewidth]{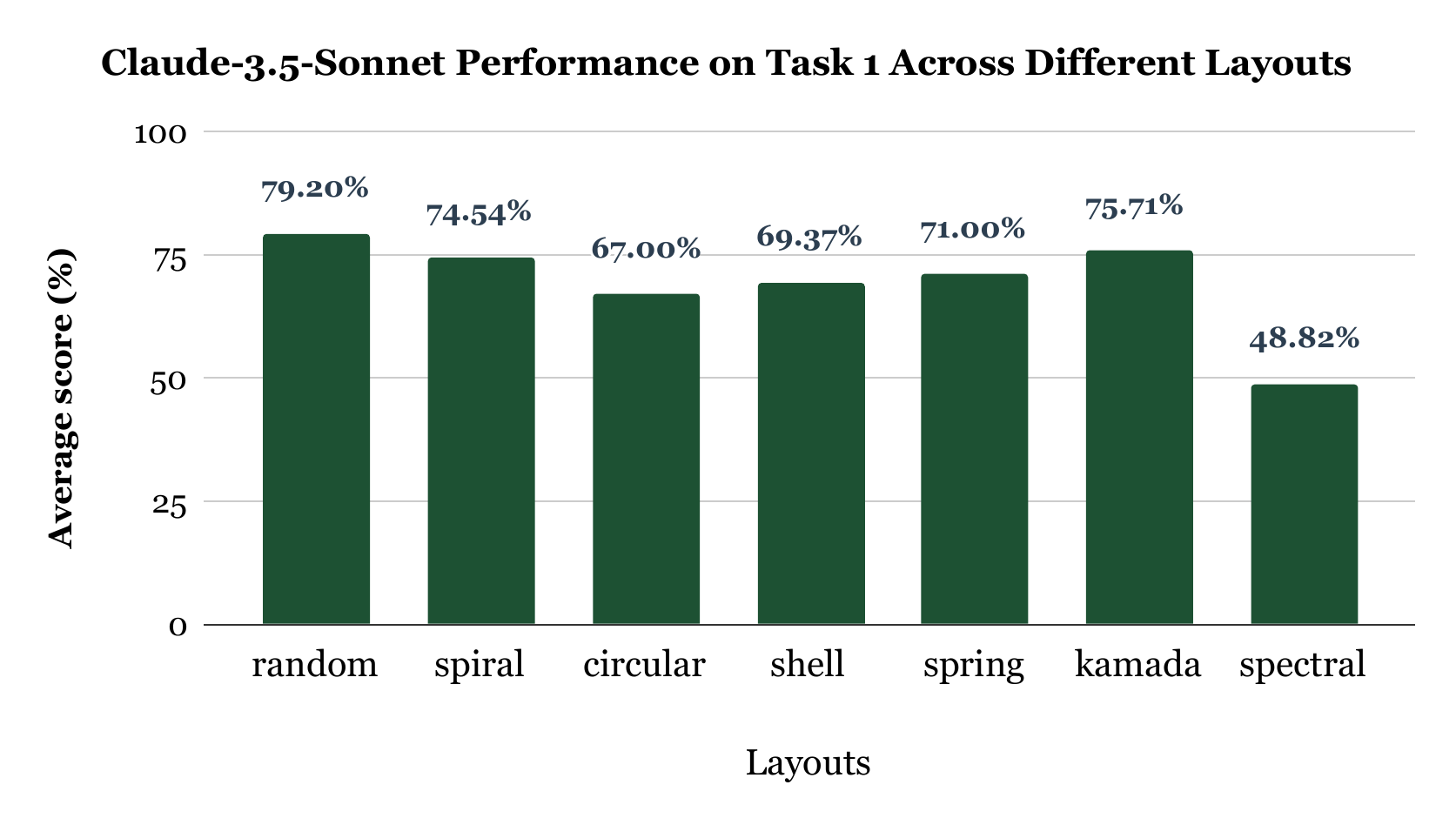}
     \caption{Average performance of Claude-3.5-Sonnet on Task 1 for each considered graph layout.}
    \label{fig:obs-spectral-layout}
\end{figure}

The bar plot in Figure~\ref{fig:obs-spectral-layout} shows the average score of the best-performing LVLM (Claude-3.5-Sonnet) on the images of Task 1 (Detection). The scores are separately shown for graphs displayed in different layouts. Interestingly, we observe a substantial decline in performance for the spectral layout (48,82\%) compared to other layouts.

\begin{figure}[t]
    \centering
    \includegraphics[width=0.6\linewidth]{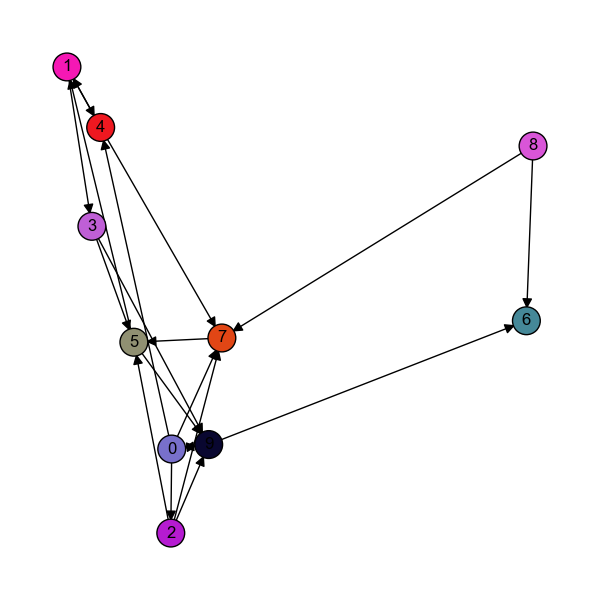}
     \caption{Image from our dataset (Task 1), showcasing a spectral layout with randomly colored nodes, directed edges, and 10 nodes with 20 edges.}
    \label{fig:layout-spectral}
\end{figure}

Figure~\ref{fig:layout-spectral} showcases such a graph with a spectral layout, consisting of 10 nodes and 20 edges. While GPT-4o accurately identifies these elements, Claude-3.5-Sonnet misidentifies the composition, reporting 9 nodes and 14 edges; see below. Interestingly, our prompts not only requested a count of elements but also encouraged the models to provide an analysis justifying their responses (see Appendix~\ref{appendix:task1}). 

\begin{tcolorbox}[enhanced,
    attach boxed title to top left={xshift=6mm,yshift=-3mm},
    colback=lightgray!50,
    colframe=lightgray,
    colbacktitle=lightgray,
    title=Output: Claude-3.5-Sonnet,
    fonttitle=\bfseries\color{black},
    boxed title style={size=small,colframe=lightgray,sharp corners},
    sharp corners,
]
\begin{minted}[breaklines, escapeinside=||]{xml}
{
   "total_nodes": |\textcolor{red}{9}|,
   "total_edges": |\textcolor{red}{14}|,
   "analysis": "The graph consists of |\textcolor{red}{9}| nodes numbered 0-8 in various colors including pink, red, purple, grey, blue, and teal. The nodes are connected by |\textcolor{red}{14}| black directed edges |\textcolor{green}{forming an asymmetric network structure. The graph has a denser cluster of connections on the left side and sparser connections extending to nodes 6 and 8 on the right.}|"
}

\end{minted}
\end{tcolorbox}

\begin{tcolorbox}[enhanced,
    attach boxed title to top left={xshift=6mm,yshift=-3mm},
    colback=lightgray!50,
    colframe=lightgray,
    colbacktitle=lightgray,
    title=Output: GPT-4o,
    fonttitle=\bfseries\color{black},
    boxed title style={size=small,colframe=lightgray,sharp corners},
    sharp corners,
]
\begin{minted}[breaklines, escapeinside=||]{xml}
```json
{
   "total_nodes": |\textcolor{green}{10}|,
   "total_edges": |\textcolor{green}{20}|,
   "analysis": "The graph contains nodes with various colors including shades of purple, red, orange, and blue. Edges are directed and have no colors."
}
```

\end{minted}
\end{tcolorbox}

As can be seen, Claude-3.5-Sonnet demonstrates a more comprehensive analysis than GPT-4o. It successfully identifies the overall node positioning, whereas GPT-4o offers a brief and irrelevant explanation. This observation suggests that when analyzing images with spectral layouts, Claude-3.5-Sonnet tends to be captivated by the layout's structural form (focusing on the analysis) rather than performing the exact count requested. This phenomenon could be attributed to the model's remarkably high performance (97.13\%) in pattern recognition (Task 4; see Section~\ref{subsec:agg-perf-eva}), which potentially biases its analytical approach towards structural interpretation over precise enumeration. Conversely, GPT-4o appears to concentrate specifically on the task at hand.

This behavior was only observed with the spectral layout, where all models worsen their predictions (see Table~\ref{tab:worse-spectral-layout}).

\begin{table}[!t]
\centering
\caption{Performance percentage for each LVLM with the spectral layout.}
\label{tab:worse-spectral-layout}
\begin{tabular}{lccc}\toprule
\multirow{2}{*}{LVLM} &\multicolumn{2}{c}{spectral layout} \\\cmidrule{2-3}
&CoT &0-shot \\\midrule
claude-3.5-sonnet &47.11 &48.82 \\
gemini-pro-1.5 &55.47 &55.11 \\
gpt-4o-2024-08-06 &64.07 &66.04 \\
llama-3.2-90b-vision-instruct &39.99 &44.34 \\
pixtral-12b &44.71 &50.92 \\
qwen-2-vl-72b-instruct &55.78 &54.56 \\
\bottomrule
\end{tabular}
\end{table}

\begin{tcolorbox}[enhanced,
    attach boxed title to top left={xshift=6mm,yshift=-3mm},
    colback=white,
    colframe=lightgreen,
    colbacktitle=lightgreen,
    title=Important,
    fonttitle=\bfseries\color{black},
    boxed title style={size=small,colframe=lightgray,sharp corners},
    sharp corners,
]
For detailed results for each layout, please visit our supplementary materials at  \url{https://camilochs.github.io/visgraphvar-website}.
\end{tcolorbox}

\subsubsection{Pixtral-12B and the Complex Task of Matching}

\begin{figure}[ht]
    \centering
    \includegraphics[width=0.8\linewidth]{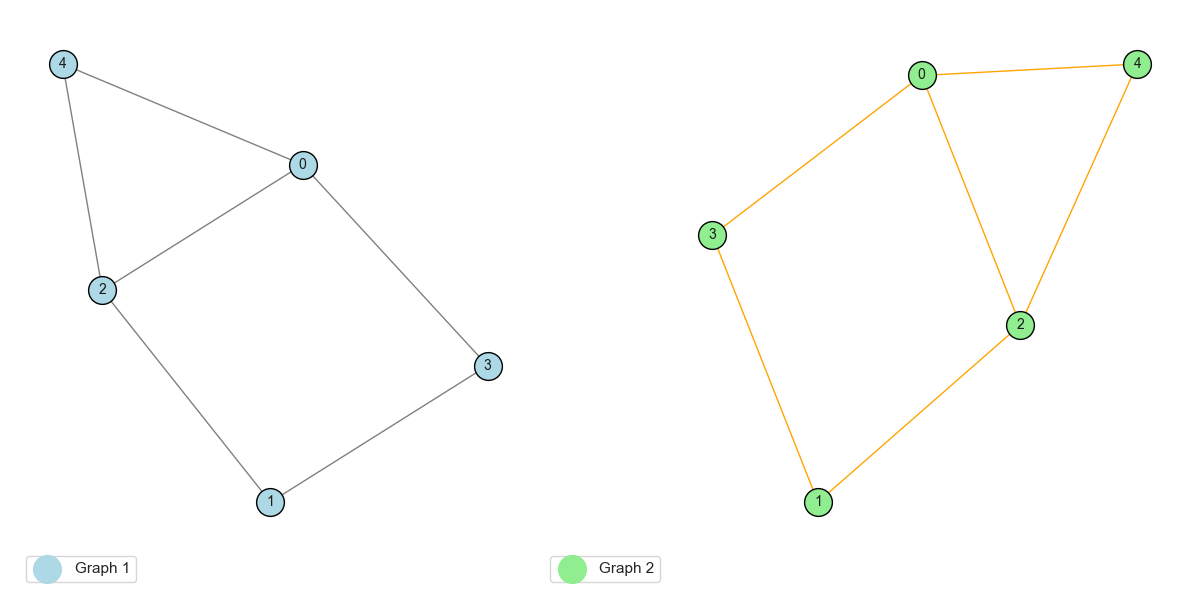}
     \caption{Image from the dataset concerning Task 7 (Matching), containing two structurally equal graphs.}
    \label{fig:fail-task7}
\end{figure}

Pixtral-12B failed in all instances of Task 7 (Matching). Remember that this dataset contains images showing graphs with 4, 5, and 6 nodes. Moreover, it was the only model to fail for all images. Figure~\ref{fig:fail-task7} shows an example of an image where it failed; see below for Pixtral-12B's response.

\begin{tcolorbox}[enhanced,
    attach boxed title to top left={xshift=6mm,yshift=-3mm},
    colback=lightgray!50,
    colframe=lightgray,
    colbacktitle=lightgray,
    title=Output: Pixtral-12B,
    fonttitle=\bfseries\color{black},
    boxed title style={size=small,colframe=lightgray,sharp corners},
    sharp corners,
]
\begin{minted}[breaklines, escapeinside=||]{xml}
{
   "match": "no",
   "analysis": "Upon examining both graphs, it is clear that they are different. The structure, the arrangement of nodes, and the connections between the nodes do not match between Graph 1 and Graph 2. |\textcolor{red}{Specifically, the filling color of the nodes and the type of lines (gray vs. orange) differ in both graphs.}|"
}
\end{minted}
\end{tcolorbox}

Pixtral-12B consistently makes errors due to a misunderstanding of the prompt (both in 0-shot and CoT scenarios). The prompts specifically ask for comparing the graphs based on their identical structure and node connections, not their visual styles (see Appendix~\ref{appendix:matching}). \visgraphvar\ presents a significant challenge with this task because it requires the model to infer that it is about graph matching, rather than comparing their stylistic similarities. Consequently, Pixtral-12B, likely due to its training set, cannot detect subtle nuances in the prompts, such as distinguishing structural equivalence from stylistic similarity in graphs.

This complexity in detecting a matching task is further highlighted by the task of detecting that graphs do not match. Paradoxically, Pixtral-12B is the only model that succeeds in all 30 cases, but for the wrong reasons. Rather than focusing on graph structure, it simply identifies a stylistic difference between Graph 1 and Graph 2 and concludes they do not match.

\subsubsection{The Impact of Node Labels on Model Performance}

\begin{figure}[t]
    \centering
    \includegraphics[width=0.9\linewidth]{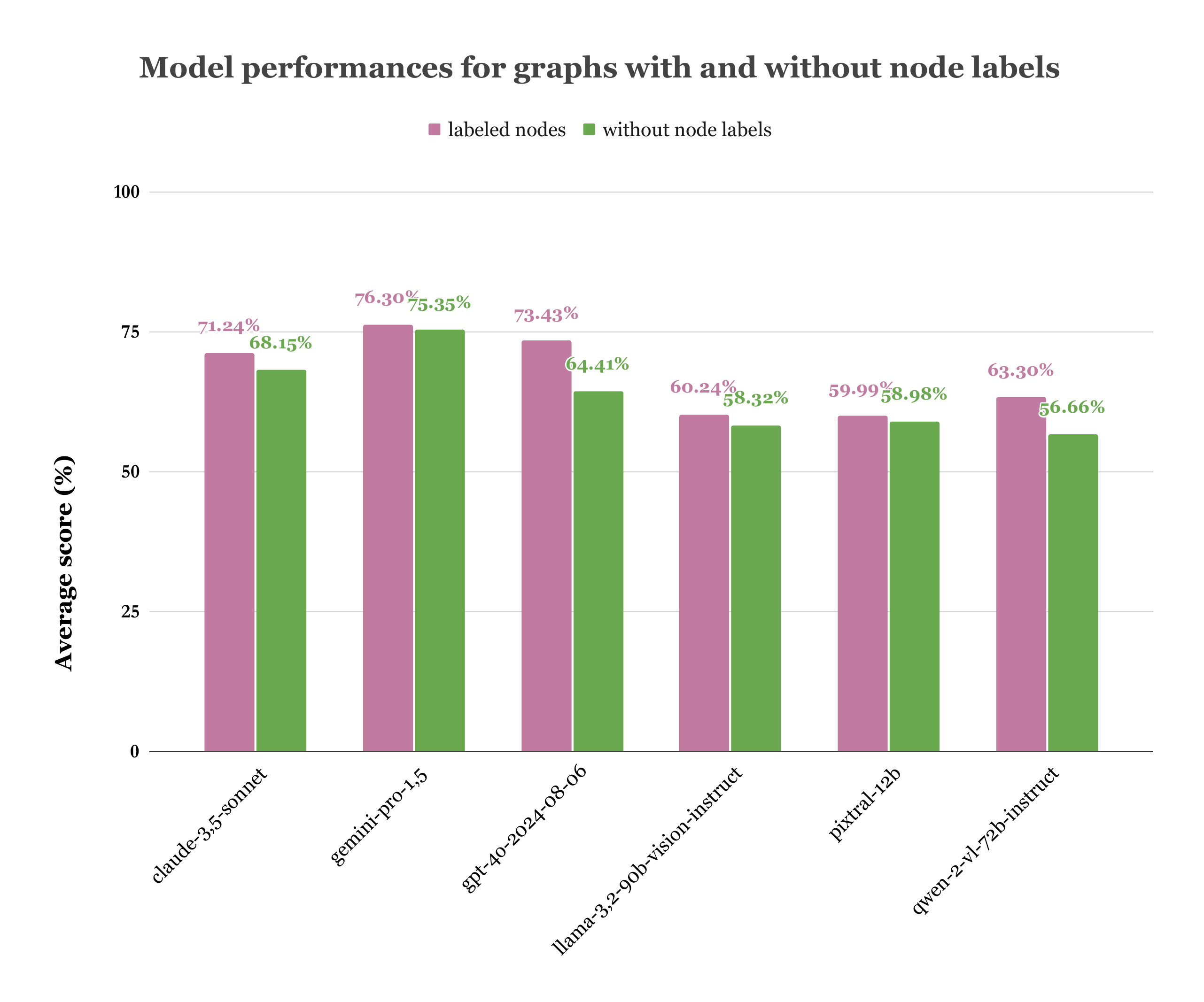}
     \caption{Comparison of the average model performance for graphs with labeled nodes (pink) and graphs with unlabeled nodes (green) in Task 1.}
    \label{fig:obs-label-task1}
\end{figure}

In the following, we present an intriguing observation that affects all models. In particular, this observation concerns possible performance differences the models show for images with node-labeled graphs and images with unlabeled graphs. Figure~\ref{fig:obs-label-task1} shows that in Task 1 (Detection), images of graphs with labeled nodes generally result in higher model performance than images with unlabeled graphs.

\begin{figure}[t]
     \centering
     \begin{subfigure}[b]{0.48\textwidth}
         \centering
         \includegraphics[width=0.9\textwidth]{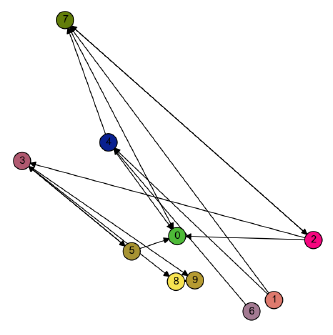}
         \caption{}
         \label{fig:obs-label-unlabel-example:1}
     \end{subfigure} 
     \begin{subfigure}[b]{0.48\textwidth}
         \centering
         \includegraphics[width=0.9\textwidth]{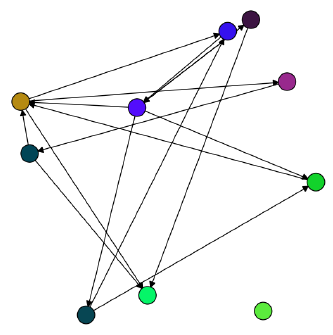}
         \caption{}
         \label{fig:obs-label-unlabel-example:2}
     \end{subfigure} 
        \caption{\textbf{(a)} A node-labeled graph with 10 nodes and 16 edges and random node-colors. \textbf{(b)} A similar, un-labeled graph. Both belong to the dataset of Task 1.}
        \label{fig:obs-label-unlabel-example}
\end{figure}

To explain this, consider the case of GPT-4o, which demonstrates the largest performance gap of $\sim9\%$. In Figure~\ref{fig:obs-label-unlabel-example}, the graphs in (a) and (b) both have 10 nodes and 16 edges. However, GPT-4o's responses produce the following discrepancies:

\begin{tcolorbox}[enhanced,
    attach boxed title to top left={xshift=6mm,yshift=-3mm},
    colback=lightgray!50,
    colframe=lightgray,
    colbacktitle=lightgray,
    title=Output: GPT-4o (for the graph in \Cref{fig:obs-label-unlabel-example:1}),
    fonttitle=\bfseries\color{black},
    boxed title style={size=small,colframe=lightgray,sharp corners},
    sharp corners,
]
\begin{minted}[breaklines, escapeinside=||]{xml}
{
   "total_nodes": |\textcolor{green}{10}|,
   "total_edges": |\textcolor{red}{18}|,
   "analysis": "The graph contains nodes of various colors, including green, pink, blue, yellow, and brown. Edges are directed and represented by black arrows connecting the nodes."
}
\end{minted}
\end{tcolorbox}

\begin{tcolorbox}[enhanced,
    attach boxed title to top left={xshift=6mm,yshift=-3mm},
    colback=lightgray!50,
    colframe=lightgray,
    colbacktitle=lightgray,
    title=Output: GPT-4o (for the graph in \Cref{fig:obs-label-unlabel-example:2}),
    fonttitle=\bfseries\color{black},
    boxed title style={size=small,colframe=lightgray,sharp corners},
    sharp corners,
]
\begin{minted}[breaklines, escapeinside=||]{xml}
{
   "total_nodes": |\textcolor{red}{9}|,
   "total_edges": |\textcolor{red}{23}|,
   "analysis": "The graph contains nodes of various colors, including green, blue, and purple, connected by directed edges. |\textcolor{green}{One node is isolated and not connected by any edges.}|"
}
\end{minted}
\end{tcolorbox}

When confronted with images of node-labeled graphs, GPT-4o accurately identifies the total number of nodes (10) and closely estimates the number of edges (18 vs. the actual 16). However, the total edges are drastically overestimated for images with unlabeled graphs, reporting 23 instead of 16 (resulting in a difference of 7). This is not an isolated case, which is shown in Figure~\ref{fig:obs-label-task1}. Although the difference is not substantial, images of graphs with labeled nodes consistently yield better performance across all models. 

We hypothesize that models with poorer performance on unlabeled graphs likely had insufficient exposure to such graphs during training, particularly in analysis tasks. This suggests an opportunity to fine-tune a model using more unlabeled graph images to evaluate whether performance improves in these cases.

%We suspect that the model's analysis may become ``skewed'' when node labels are present, as it potentially places less emphasis on accurately detecting edges. Conversely, in the absence of node labels, the model has fewer visual elements to “distract” its attention, allowing it to focus more effectively on the graph's structural details. This is further supported by the analysis of the unlabeled node graph, as it correctly identifies the isolated node.

\section{Discussion and Open Questions}\label{sec:discussion}

The multimodal capabilities of LLMs have greatly expanded their potential, with LVLMs standing out as a prime example. With just an API and at a low cost, one can now obtain detailed image analyses, whether for object detection or generating comprehensive image captions. However, tasks that involve vision and inference remain inherently difficult and have a long history in computer vision. To properly evaluate emerging LVLMs, they must be tested against visually complex tasks. This is where flexible geometric structures like graphs become crucial. Graphs can introduce significant complexity by simply altering the spatial arrangement of nodes and edges, or modifying their visual representation.

Our study introduced \visgraphvar, a benchmark generator designed to challenge current LVLMs and to serve as a testing platform for future models focused on visual graph inference. Below, we present open questions from our study that warrant further exploration:

\begin{enumerate}
    \item \textbf{Which visual style changes truly impair a model's prediction?} Through the dataset of graph images produced with \visgraphvar, we provided evidence showing that visual changes affect inference performance, for example, by adding node labels or modifying the layout. However, identifying exactly which changes harm model performance and quantifying their impact on results remains an open question. We trust \visgraphvar\ can inspire other researchers to delve deeper into these questions---not just for visual graph inference---by conducting comprehensive analyses to clearly identify which visual styles distort predictions.

    \item \textbf{Can an LVLM achieve an accuracy score of over 90\% across all seven tasks covered by \visgraphvar?} Testing an LVLM against these seven tasks proved challenging, and although Claude-3.5-Sonnet achieved the best results, it is still far from attaining high scores across all tasks. Some ideas to improve results on our dataset, or future datasets generated with \visgraphvar, include the following ones:
    \begin{itemize}
        \item Training a fine-tuned model that can cope with the seven tasks and variations in visual styles~\cite{zhai2024finetuninglargevisionlanguagemodels}.
        \item Researching prompt strategies~\cite{vatsal2024surveypromptengineeringmethods,sahoo2024systematicsurveypromptengineering} and hypothesizing whether prompt modifications could improve results.
        \item Avoiding single prompts in favor of a series of prompts that apply self-correction in the context of an image~\cite{Pan2023-ll}, generating interaction through LLM-based agents~\cite{renze2024selfreflectionllmagentseffects}.
    \end{itemize}

    \item \textbf{How can LVLMs be incorporated into real-world graph theory applications?} VisionGraph~\cite{li2024visiongraphleveraginglargemultimodal}, GITA~\cite{wei2024gitagraphvisualtextual}, and our benchmark generator \visgraphvar\ focus on small graphs, as LVLMs are not yet mature enough to work at a larger scale. Furthermore, when dealing with larger graphs (for example, those with high density), the visualization challenges are not unique to LVLMs but rather a common limitation faced by any automated interpretation system working with such complex graph structures. However, we can anticipate that once they become more precise in various graph analysis tasks, they will be helpful in quickly analyzing complex graph images, for example, in social networks, communication networks, and education applications. We consider that with the current state of multimodal model technology, it is already possible to incorporate them into applications in the case of general analysis questions rather than precise algorithmic operations on an image. To integrate an LVLM, the first step would be to define which tasks it should perform (using the present study as a guide) and then determine which model performs best for those tasks. 
\end{enumerate}

%\subsection{Implications for Graph Theory}
%\subsection{Applications in Various Domains}
%\subsection{Future Research Directions}
%\subsection{Advancing LVLMs with Prompt Engineering}

\section{Conclusions}\label{sec:conclu}

This research evaluated Large Vision-Language Models (LVLMs) through the lens of graphs, chosen for their rich geometric properties and structural complexity. We introduced \visgraphvar, a customizable benchmark generator designed to explore visual variability across both stylistic elements and graph structures. The framework generates datasets spanning seven distinct tasks (detection, classification, segmentation, pattern recognition, link prediction, reasoning, and matching), providing comprehensive coverage of graph interpretation and analysis capabilities. Our analysis of the \visgraphvar-generated dataset reveals two key findings: first, that visual variability has a substantial impact on LVLM performance, and second, that current LVLM architectures require fundamental enhancements to adequately address the complexities of visual graph interpretation tasks.

\subsection{Future Work}

 \visgraphvar's inherent flexibility creates opportunities for future research, such as generating increasingly complex graph image datasets with more advanced node and edge configurations. This would enable more detailed evaluations, offering deeper insights into the capabilities and limitations of LVLMs in visual graph analysis tasks.

\section*{Acknowledgements}
The research presented in this paper was supported by grants TED2021-129319B-I00 and PID2022-136787NB-I00 funded by MCIN/AEI/10.13039/ 501100011033.
%%

%Bibliography
\bibliographystyle{plainnat}  
\bibliography{references}  

\begin{appendices}

\section{Prompt design}\label{app:prompt-design}

Our evaluation methodology for the dataset generated by \visgraphvar\ required the development of 14 distinct prompts, comprising two prompting strategies (zero-shot and chain-of-thought) across seven tasks. We automated this process by first hand-crafting a seed prompt that instructed Claude-3.5-Sonnet to generate prompts for each strategy. Through iterative refinement of this approach, we developed the subsequent six prompt pairs. This systematic process resulted in our final set of 14 prompts.

Below, we present our initial seed prompt for Task 1, accompanied by the two corresponding prompts generated by Claude-3.5-Sonnet. The complete collection of prompts is available in the supplementary material.

\begin{tcolorbox}[enhanced,
    attach boxed title to top left={xshift=6mm,yshift=-3mm},
    colback=white,
    colframe=lightgray,
    colbacktitle=lightgray,
    title=Prompt written by us (Task 1),
    fonttitle=\bfseries\color{black},
    boxed title style={size=small,colframe=lightgray,sharp corners},
    sharp corners,
]
\begin{minted}[breaklines, escapeinside=||]{xml}
Generate two prompts for graph analysis using a Large Vision Model.

1. Zero-shot prompt: Ask the model to analyze the graph image without any prior examples or hints. The goal is to answer the question: How many nodes and edges are in the image?

2. Chain-of-thought prompt: Ask the model to reason step by step through the process of identifying the nodes and edges in the graph. The question remains: How many nodes and edges are in the image?

Expected output format (JSON example):
```json
{
   "total_nodes": 10,
   "total_edges": 5,
   "analysis": "The image shows ten nodes with color red and five edges with color black."
}
```
\end{minted}
\end{tcolorbox}

\begin{tcolorbox}[enhanced,
    attach boxed title to top left={xshift=6mm,yshift=-3mm},
    colback=lightgray!50,
    colframe=lightgray,
    colbacktitle=lightgray,
    title=Zero-shot Prompt Output Generated by Claude-3.5-Sonnet,
    fonttitle=\bfseries\color{black},
    boxed title style={size=small,colframe=lightgray,sharp corners},
    sharp corners,
]
\begin{minted}[breaklines, escapeinside=||]{xml}
Analyze the graph in this image. Count and report the number of nodes and edges present. Provide your answer in the following JSON format:

```json
{
   "total_nodes": [number of nodes],
   "total_edges": [number of edges],
   "analysis": "[Brief description of the graph, including node and edge colors]"
}
```
\end{minted}
\end{tcolorbox}

\begin{tcolorbox}[enhanced,
    attach boxed title to top left={xshift=6mm,yshift=-3mm},
    colback=lightgray!50,
    colframe=lightgray,
    colbacktitle=lightgray,
    title=Chain-of-Thought Prompt Output Generated by Claude-3.5-Sonnet,
    fonttitle=\bfseries\color{black},
    boxed title style={size=small,colframe=lightgray,sharp corners},
    sharp corners,
]
\begin{minted}[breaklines, escapeinside=||]{xml}
Let's analyze the graph in this image step by step to determine the number of nodes and edges. Follow these steps:

1. Identify the nodes:
   - Look for distinct points or circles in the graph.
   - Count each unique node you see.
   - Note the color of the nodes.

2. Identify the edges:
   - Look for lines connecting the nodes.
   - Count each unique edge you see.
   - Note the color of the edges.

3. Summarize your findings:
   - Report the total number of nodes.
   - Report the total number of edges.
   - Briefly describe the graph's appearance.

4. Format your answer:
   Present your analysis in the following JSON format:

   ```json
   {
      "total_nodes": [number of nodes],
      "total_edges": [number of edges],
      "analysis": "[Brief description of the graph, including node and edge colors]"
   }
   ```

\end{minted}
\end{tcolorbox}

\begin{comment}
    
\end{comment}
\section{Input-Output Examples}\label{secA1}

Below are representative examples of each task covered by \visgraphvar, including zero-shot prompts and contrasting results from two models to show both successful and unsuccessful outcomes. Complete results and prompts for all evaluated models can be found in the supplementary material. 
\subsection{Task 1: node and edge detection}\label{appendix:task1}
\begin{tcolorbox}[
    enhanced,
    attach boxed title to top left={xshift=6mm,yshift=-3mm},
    colback=white,
    colframe=lightgray,
    colbacktitle=lightgray,
    title=Input: 0-shot prompt (left) and image (right),
    fonttitle=\bfseries\color{black},
    boxed title style={size=small,colframe=lightgray,sharp corners},
    sharp corners,
]

\begin{minipage}[t]{0.5\linewidth}
    \vspace*{0pt}

\begin{minted}[breaklines]{xml}
Analyze the graph in this image. Count and report the number of nodes and edges present. Provide your answer in the following JSON format:

```json
{
   "total_nodes": [number of nodes],
   "total_edges": [number of edges],
   "analysis": "[Brief description of the graph, including node and edge colors]"
}
```
\end{minted}
    
    \end{minipage}
    \begin{minipage}[t]{0.5\linewidth}
    \vspace*{1pt}
        \includegraphics[height=0.28\textheight,width=\linewidth]{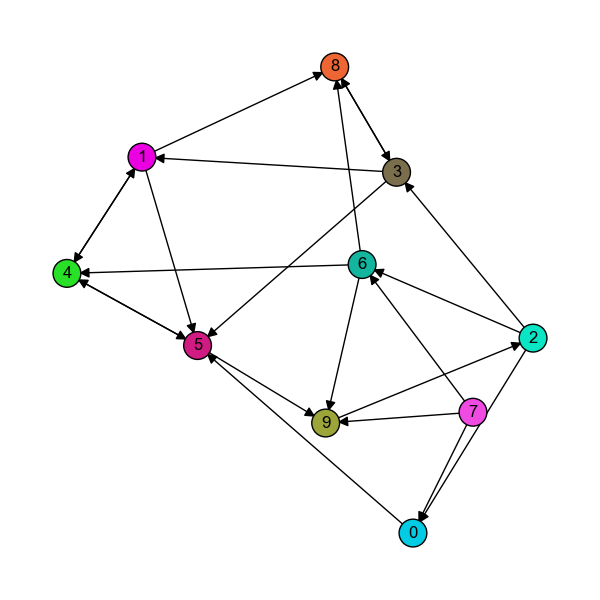}
       \label{fig:fig1}
    \end{minipage}

\end{tcolorbox}
\begin{tcolorbox}[enhanced,
    attach boxed title to top left={xshift=6mm,yshift=-3mm},
    colback=lightgray!50,
    colframe=lightgray,
    colbacktitle=lightgray,
    title=Output: gemini-pro-1.5,
    fonttitle=\bfseries\color{black},
    boxed title style={size=small,colframe=lightgray,sharp corners},
    sharp corners,
]
\begin{minted}[breaklines, escapeinside=||]{xml}
{
  "total_nodes": |\textcolor{green}{10}|,
  "total_edges": |\textcolor{green}{22}|,
  "analysis": "The graph has 10 nodes, numbered from 0 to 9, and 22 directed edges. Each node has a distinct color: Node 0 is cyan, 1 is magenta, 2 is turquoise, 3 is olive, 4 is lime green, 5 is medium violet red, 6 is light sea green, 7 is hot pink, 8 is chocolate, and 9 is olive drab."
}
\end{minted}
\end{tcolorbox}

\begin{tcolorbox}[enhanced,
    attach boxed title to top left={xshift=6mm,yshift=-3mm},
    colback=lightgray!50,
    colframe=lightgray,
    colbacktitle=lightgray,
    title=Output: pixtral-12b,
    fonttitle=\bfseries\color{black},
    boxed title style={size=small,colframe=lightgray,sharp corners},
    sharp corners,
]
\begin{minted}[breaklines, escapeinside=||]{xml}
{
   "total_nodes": |\textcolor{red}{9}|,
   "total_edges": |\textcolor{red}{23}|,
   "analysis": "This graph contains 9 colored nodes connected by 23 directed edges. The nodes are colored with different colors: orange, magenta, green, cyan, brown, olive, and yellow. This implies that the nodes are likely categorized or classified differently, possibly representing distinct elements or entities in a network or system with multiple interrelations."
}
\end{minted}
\end{tcolorbox}

\subsection{Task 2: classification}
\begin{tcolorbox}[
    enhanced,
    attach boxed title to top left={xshift=6mm,yshift=-3mm},
    colback=white,
    colframe=lightgray,
    colbacktitle=lightgray,
    title=Input: 0-shot prompt (left) and image (right),
    fonttitle=\bfseries\color{black},
    boxed title style={size=small,colframe=lightgray,sharp corners},
    sharp corners,
]

\begin{minipage}[t]{0.5\linewidth}
    \vspace*{0pt}

\begin{minted}[breaklines]{xml}
Please analyze this graph visualization and classify it into one of the following categories: acyclic, bipartite, complete, cyclic, mesh, planar, or tree.

Provide your response in the following JSON format:

```json
{
   "type_graph": "[classification]",
   "analysis": "The image shows a [classification] graph."
}
```
\end{minted}
    
    \end{minipage}
    \begin{minipage}[t]{0.5\linewidth}
    \vspace*{1pt}
        \includegraphics[height=0.28\textheight,width=\linewidth]{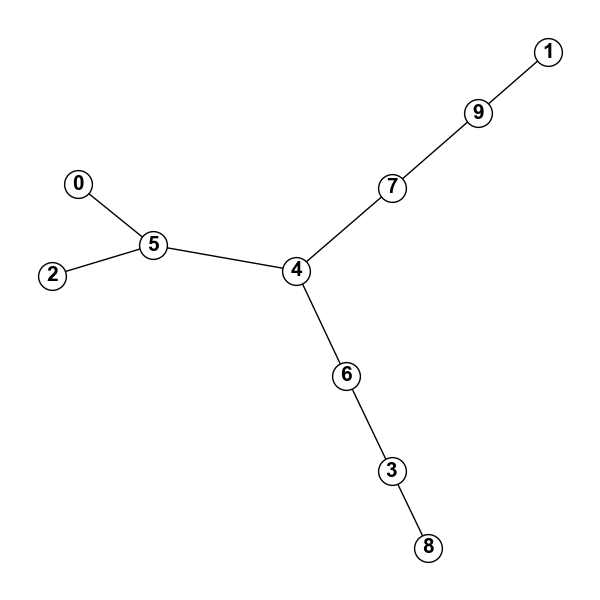}
       \label{fig:fig2}
    \end{minipage}

\end{tcolorbox}
\begin{tcolorbox}[enhanced,
    attach boxed title to top left={xshift=6mm,yshift=-3mm},
    colback=lightgray!50,
    colframe=lightgray,
    colbacktitle=lightgray,
    title=Output: claude-3.5-Sonnet,
    fonttitle=\bfseries\color{black},
    boxed title style={size=small,colframe=lightgray,sharp corners},
    sharp corners,
]
\begin{minted}[breaklines, escapeinside=||]{xml}
{
   "type_graph": |\textcolor{green}{"tree"}|,
   "analysis": "The image shows a tree graph, as it is connected, has no cycles, and each node is connected by exactly one path. The graph has 10 nodes (labeled 0-9) with edges forming a branching structure without any loops."
}
\end{minted}
\end{tcolorbox}

\begin{tcolorbox}[enhanced,
    attach boxed title to top left={xshift=6mm,yshift=-3mm},
    colback=lightgray!50,
    colframe=lightgray,
    colbacktitle=lightgray,
    title=Output: qwen-2-vl-72b-instruct,
    fonttitle=\bfseries\color{black},
    boxed title style={size=small,colframe=lightgray,sharp corners},
    sharp corners,
]
\begin{minted}[breaklines, escapeinside=||]{xml}
{
   "type_graph": |\textcolor{red}{"acyclic"}|,
   "analysis": "The image shows a acyclic graph."
}
\end{minted}
\end{tcolorbox}

\subsection{Task 3: segmentation}
\begin{tcolorbox}[
    enhanced,
    attach boxed title to top left={xshift=6mm,yshift=-3mm},
    colback=white,
    colframe=lightgray,
    colbacktitle=lightgray,
    title=Input: 0-shot prompt (left) and image (right),
    fonttitle=\bfseries\color{black},
    boxed title style={size=small,colframe=lightgray,sharp corners},
    sharp corners,
]

\begin{minipage}[t]{0.5\linewidth}
    \vspace*{0pt}

\begin{minted}[breaklines]{xml}
Analyze this graph visualization and identify the cut edge that, when removed, divides the graph into two separate subgraphs. A cut edge (or bridge) is a single edge whose removal increases the number of connected components.

Provide your response in the following JSON format:
```json
{
   "cut_edge": "(node1, node2)",
   "analysis": "Removing the edge between nodes [node1] and [node2] splits the graph into two subgraphs: [brief description of resulting subgraphs]."
}
```
\end{minted}
    
    \end{minipage}
    \begin{minipage}[t]{0.5\linewidth}
    \vspace*{1pt}
        \includegraphics[height=0.28\textheight,width=\linewidth]{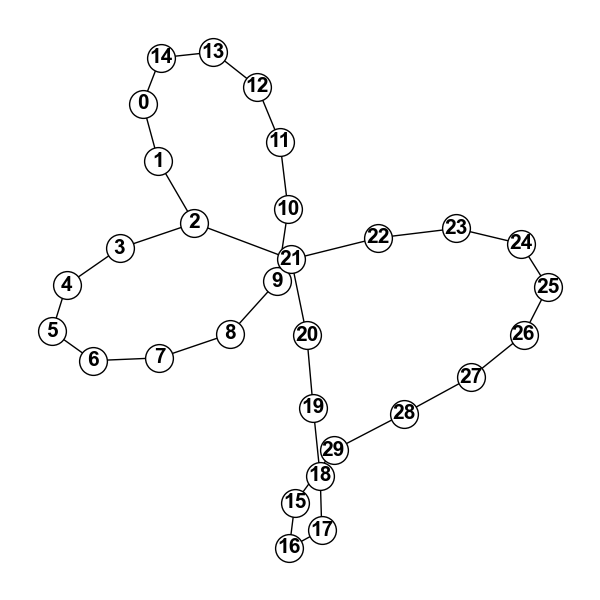}
       \label{fig:fig3}
    \end{minipage}

\end{tcolorbox}
\begin{tcolorbox}[enhanced,
    attach boxed title to top left={xshift=6mm,yshift=-3mm},
    colback=lightgray!50,
    colframe=lightgray,
    colbacktitle=lightgray,
    title=Output: gemini-pro-1.5,
    fonttitle=\bfseries\color{black},
    boxed title style={size=small,colframe=lightgray,sharp corners},
    sharp corners,
]
\begin{minted}[breaklines, escapeinside=||]{xml}
{
  "cut_edge": "(|\textcolor{green}{21}|, |\textcolor{red}{22}|)",
  "analysis": "Removing the edge between nodes 21 and 22 splits the graph into two subgraphs: One subgraph contains nodes 0-20 and the other subgraph contains nodes 22-29."
}
\end{minted}
\end{tcolorbox}

\begin{tcolorbox}[enhanced,
    attach boxed title to top left={xshift=6mm,yshift=-3mm},
    colback=lightgray!50,
    colframe=lightgray,
    colbacktitle=lightgray,
    title=Output: qwen-2-vl-72b-instruct,
    fonttitle=\bfseries\color{black},
    boxed title style={size=small,colframe=lightgray,sharp corners},
    sharp corners,
]
\begin{minted}[breaklines, escapeinside=||]{xml}
{
   "cut_edge": "(|\textcolor{red}{9}|, |\textcolor{red}{20}|)",
   "analysis": "Removing the edge between nodes [9] and [20] splits the graph into two subgraphs: [a subgraph containing nodes 0-8 (inclusive)] and [a subgraph containing nodes 9-29 (inclusive)]."
}
\end{minted}
\end{tcolorbox}

\subsection{Task 4: pattern recognition}
\begin{tcolorbox}[
    enhanced,
    attach boxed title to top left={xshift=6mm,yshift=-3mm},
    colback=white,
    colframe=lightgray,
    colbacktitle=lightgray,
    title=Input: 0-shot prompt (left) and image (right),
    fonttitle=\bfseries\color{black},
    boxed title style={size=small,colframe=lightgray,sharp corners},
    sharp corners,
]

\begin{minipage}[t]{0.5\linewidth}
    \vspace*{0pt}

\begin{minted}[breaklines]{xml}
Analyze the following graph image and identify the type of pattern it contains. The possible patterns are 'chain,' 'clique,' or 'star.' Please respond in the JSON format provided:

```json
{
   'pattern': [name],
   'number_of_patterns': [number],
   'analysis': '[explain].'
}
```
Where:

    - pattern is the name of the pattern type you detect.
    - number_of_patterns is the total count of this pattern type found.
    - analysis is a brief explanation of the features in the graph that led you to your conclusion.
\end{minted}
    
    \end{minipage}
    \begin{minipage}[t]{0.5\linewidth}
    \vspace*{1pt}
        \includegraphics[height=0.28\textheight,width=\linewidth]{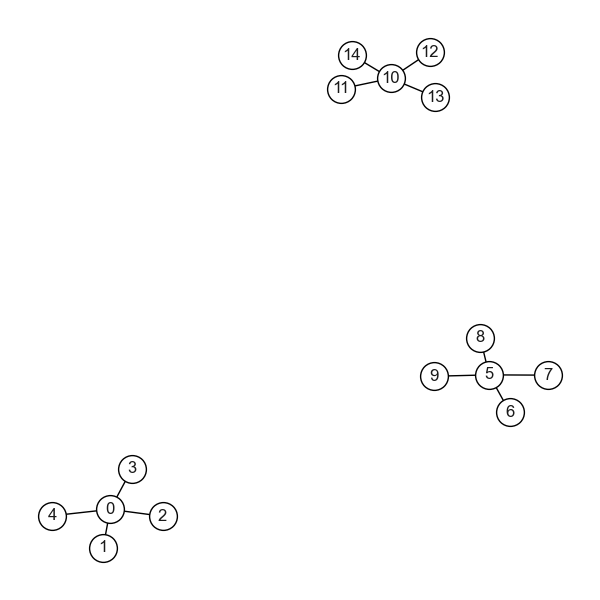}
       \label{fig:fig4}
    \end{minipage}

\end{tcolorbox}
\begin{tcolorbox}[enhanced,
    attach boxed title to top left={xshift=6mm,yshift=-3mm},
    colback=lightgray!50,
    colframe=lightgray,
    colbacktitle=lightgray,
    title=Output: gemini-pro-1.5,
    fonttitle=\bfseries\color{black},
    boxed title style={size=small,colframe=lightgray,sharp corners},
    sharp corners,
]
\begin{minted}[breaklines, escapeinside=||]{xml}
{
    "pattern": |\textcolor{green}{"star"}|,
    "number_of_patterns": |\textcolor{green}{3}|,
    "analysis": "The graph contains three distinct star patterns. Each star has a central node connected to multiple peripheral nodes. The first star has node 0 as center connecting to nodes 1,2,3,4. The second star has node 5 as center connecting to nodes 6,7,8,9. The third star has node 10 as center connecting to nodes 11,12,13,14. Each central node acts as a hub with radiating connections, which is characteristic of a star pattern."
}
\end{minted}
\end{tcolorbox}

\begin{tcolorbox}[enhanced,
    attach boxed title to top left={xshift=6mm,yshift=-3mm},
    colback=lightgray!50,
    colframe=lightgray,
    colbacktitle=lightgray,
    title=Output: qwen-2-vl-72b-instruct,
    fonttitle=\bfseries\color{black},
    boxed title style={size=small,colframe=lightgray,sharp corners},
    sharp corners,
]
\begin{minted}[breaklines, escapeinside=||]{xml}
{
   "pattern": |\textcolor{green}{"star"}|,
   "number_of_patterns": |\textcolor{red}{2}|,
   "analysis": "The graph contains two star patterns. One centered on node 0 with leaves 1, 2, 3, and 4. The other is centered on node 5 with leaves 6, 7, 8, and 9. Both patterns have a central node connected to several others, similar to a 'star' configuration."
}
\end{minted}
\end{tcolorbox}

\subsection{Task 5: link prediction}
\begin{tcolorbox}[
    enhanced,
    attach boxed title to top left={xshift=6mm,yshift=-3mm},
    colback=white,
    colframe=lightgray,
    colbacktitle=lightgray,
    title=Input: 0-shot prompt (left) and image (right),
    fonttitle=\bfseries\color{black},
    boxed title style={size=small,colframe=lightgray,sharp corners},
    sharp corners,
]

\begin{minipage}[t]{0.5\linewidth}
    \vspace*{0pt}

\begin{minted}[breaklines]{xml}
Given the graph image before you, analyze its structure and topology. Based on the existing connection patterns and node relationships, identify two currently unconnected nodes that should logically be connected with an edge. Consider the graph's natural organizational principles and structural patterns.

Your response should be formatted as a JSON object with the following structure:
```json
{
    "nodes_prediction": [node1, node2],
    "analysis": "Brief explanation of why these nodes should be connected based on the graph's structure."
}
```
\end{minted}
    
    \end{minipage}
    \begin{minipage}[t]{0.5\linewidth}
    \vspace*{1pt}
        \includegraphics[height=0.28\textheight,width=\linewidth]{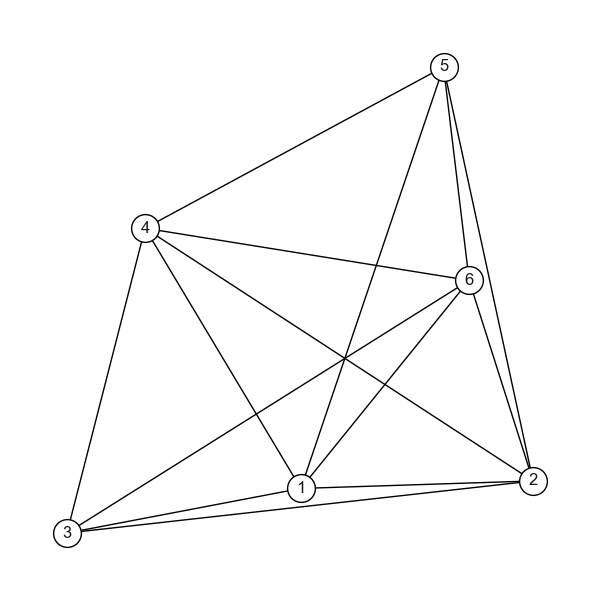}
       \label{fig:fig5}
    \end{minipage}

\end{tcolorbox}
\begin{tcolorbox}[enhanced,
    attach boxed title to top left={xshift=6mm,yshift=-3mm},
    colback=lightgray!50,
    colframe=lightgray,
    colbacktitle=lightgray,
    title=Output: gemini-pro-1.5,
    fonttitle=\bfseries\color{black},
    boxed title style={size=small,colframe=lightgray,sharp corners},
    sharp corners,
]
\begin{minted}[breaklines, escapeinside=||]{xml}
{
  "nodes_prediction": [|\textcolor{green}{3}|, |\textcolor{green}{5}|],
  "analysis": "Looking at the graph structure, nodes 3 and 5 should logically be connected. This prediction is based on several observations: (1) The graph shows a pattern of triangular formations and high connectivity between most nodes, (2) Most nodes have connections to at least 4 other nodes, creating a relatively dense network, (3) Nodes 3 and 5 are both connected to nodes 4 and 6, suggesting they are part of the same cluster or community within the graph. Adding an edge between nodes 3 and 5 would complete another triangular formation and maintain the graph's overall structural symmetry and connectivity patterns."
}
\end{minted}
\end{tcolorbox}

\begin{tcolorbox}[enhanced,
    attach boxed title to top left={xshift=6mm,yshift=-3mm},
    colback=lightgray!50,
    colframe=lightgray,
    colbacktitle=lightgray,
    title=Output: pixtral-12b,
    fonttitle=\bfseries\color{black},
    boxed title style={size=small,colframe=lightgray,sharp corners},
    sharp corners,
]
\begin{minted}[breaklines, escapeinside=||]{xml}
{
    "nodes_prediction": [|\textcolor{green}{3}|, |\textcolor{red}{6}|],
    "analysis": "Based on the graph's structure and the existing connections, it seems that node 3 and node 6 are both part of the interconnected clusters of the graph. Connecting them would enhance the connectivity and likely maintain a balanced structure similar to the existing connections within the graph."
}
\end{minted}
\end{tcolorbox}

\subsection{Task 6: reasoning}
\begin{tcolorbox}[
    enhanced,
    attach boxed title to top left={xshift=6mm,yshift=-3mm},
    colback=white,
    colframe=lightgray,
    colbacktitle=lightgray,
    title=Input: 0-shot prompt (left) and image (right),
    fonttitle=\bfseries\color{black},
    boxed title style={size=small,colframe=lightgray,sharp corners},
    sharp corners,
]

\begin{minipage}[t]{0.5\linewidth}
    \vspace*{0pt}

\begin{minted}[breaklines]{xml}
Analyze the given graph image and identify the shortest path between the nodes labeled 1 and 6. Consider only the visible connections and any weighted or unweighted paths. Return the answer in JSON format, detailing the nodes along the path and a brief explanation.

Expected output format:
```json
{
   "shortest_path": ["1", "...", "6"],
   "analysis": "Explanation of path selection and any notable features in the graph structure."
}
```
\end{minted}
    
    \end{minipage}
    \begin{minipage}[t]{0.5\linewidth}
    \vspace*{1pt}
        \includegraphics[height=0.28\textheight,width=\linewidth]{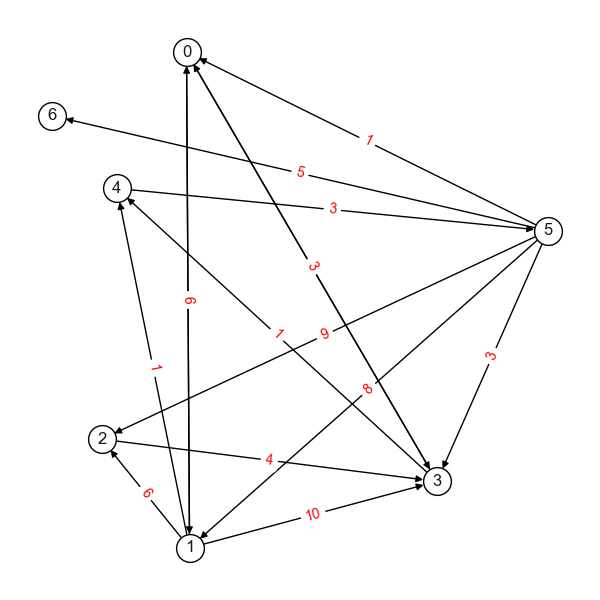}
       \label{fig:fig6}
    \end{minipage}

\end{tcolorbox}
\begin{tcolorbox}[enhanced,
    attach boxed title to top left={xshift=6mm,yshift=-3mm},
    colback=lightgray!50,
    colframe=lightgray,
    colbacktitle=lightgray,
    title=Output: gemini-pro-1.5,
    fonttitle=\bfseries\color{black},
    boxed title style={size=small,colframe=lightgray,sharp corners},
    sharp corners,
]
\begin{minted}[breaklines, escapeinside=||]{xml}
{
  "shortest_path": [|\textcolor{green}{"1"}|, |\textcolor{red}{"2"}|, |\textcolor{red}{"4"}|, |\textcolor{green}{"6"}|],
  "analysis": "The shortest path from node 1 to node 6 involves traversing through nodes 2 and 4. The edge weights along this path (1, 1) sum to 2, which represents a shorter distance than any other path involving additional nodes or edges with higher weights like the path through 1 -> 0 adding up to 7 or 1 -> 3 -> 5 -> 6 adding up to 14."
}
\end{minted}
\end{tcolorbox}

\begin{tcolorbox}[enhanced,
    attach boxed title to top left={xshift=6mm,yshift=-3mm},
    colback=lightgray!50,
    colframe=lightgray,
    colbacktitle=lightgray,
    title=Output: qwen-2-vl-72b-instruct,
    fonttitle=\bfseries\color{black},
    boxed title style={size=small,colframe=lightgray,sharp corners},
    sharp corners,
]
\begin{minted}[breaklines, escapeinside=||]{xml}
{
   "shortest_path": [|\textcolor{green}{"1"}|, |\textcolor{red}{"2"}|, |\textcolor{red}{"0"}|, |\textcolor{green}{"5"}|, |\textcolor{green}{"6"}|],
   "analysis": "The shortest path from node 1 to node 6 can be determined by following the edges with the smallest cumulative weight. Starting from node 1, the path goes through node 2 (weight 4), then to node 0 (weight 2), next to node 5 (weight 5), and finally to node 6 (weight 1). This path has a total weight of 12, which is the smallest possible sum of weights for any path from node 1 to node 6. Notable features include the direct connections and the varying weights of the edges, which influence the selection of the shortest path."
}
\end{minted}
\end{tcolorbox}

\subsection{Task 7: matching} \label{appendix:matching}
\begin{tcolorbox}[
    enhanced,
    attach boxed title to top left={xshift=6mm,yshift=-3mm},
    colback=white,
    colframe=lightgray,
    colbacktitle=lightgray,
    title=Input: 0-shot prompt (left) and image (right),
    fonttitle=\bfseries\color{black},
    boxed title style={size=small,colframe=lightgray,sharp corners},
    sharp corners,
]

\begin{minipage}[t]{0.5\linewidth}
    \vspace*{0pt}

\begin{minted}[breaklines]{xml}
Examine the two given graph images and determine if they match. Matching implies identical structures, node connections, and any weights if present. Provide a response in JSON format with a 'yes' or 'no' answer and a brief explanation.

Expected output format:
```json
{
   "match": "yes" or "no",
   "analysis": "Explanation of key similarities or differences observed between the graphs."
}
```
\end{minted}
    
    \end{minipage}
    \begin{minipage}[t]{0.5\linewidth}
    \vspace*{1pt}
        \includegraphics[height=0.18\textheight,width=\linewidth]{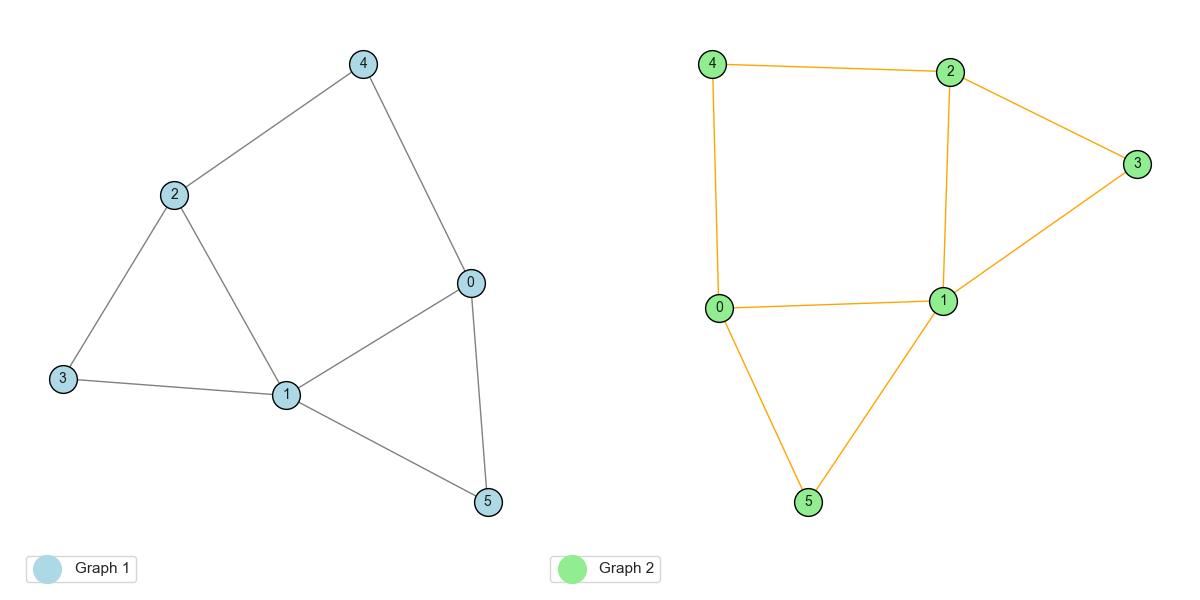}
       \label{fig:fig7}
    \end{minipage}

\end{tcolorbox}
\begin{tcolorbox}[enhanced,
    attach boxed title to top left={xshift=6mm,yshift=-3mm},
    colback=lightgray!50,
    colframe=lightgray,
    colbacktitle=lightgray,
    title=Output: gemini-pro-1.5,
    fonttitle=\bfseries\color{black},
    boxed title style={size=small,colframe=lightgray,sharp corners},
    sharp corners,
]
\begin{minted}[breaklines, escapeinside=||]{xml}
{
   "match": |\textcolor{green}{"yes"}|,
   "analysis": "Both graphs contain the same structure with 6 nodes (labeled 0-5) and identical connectivity patterns. While the layouts and colors appear different (blue nodes in Graph 1, green nodes in Graph 2), the underlying graph structure is the same. All node connections match between the two graphs, including the connections between nodes 0-1, 1-5, 0-4, 1-3, 2-3, 2-4, and all other edges. The graphs are isomorphic, meaning they represent the same mathematical structure despite their different visual presentations."
}
\end{minted}
\end{tcolorbox}

\begin{tcolorbox}[enhanced,
    attach boxed title to top left={xshift=6mm,yshift=-3mm},
    colback=lightgray!50,
    colframe=lightgray,
    colbacktitle=lightgray,
    title=Output: qwen-2-vl-72b-instruct,
    fonttitle=\bfseries\color{black},
    boxed title style={size=small,colframe=lightgray,sharp corners},
    sharp corners,
]
\begin{minted}[breaklines, escapeinside=||]{xml}
{
   "match": |\textcolor{red}{"no"}|,
   "analysis": "The two graphs have different structures. Graph 1 has a node labeled '4' connected to nodes '2' and '0', while in Graph 2, the node labeled '4' is connected to nodes '2' and '0' with a different layout. Additionally, the connections between nodes in Graph 1 and Graph 2 do not match. For example, in Graph 1, node '1' is connected to nodes '0', '2', and '3', whereas in Graph 2, node '1' is connected to nodes '0', '5', and '3'."
}
\end{minted}
\end{tcolorbox}




\end{appendices}

\end{document}